\PassOptionsToPackage{dvipsnames,table}{xcolor}
\documentclass[acmsmall]{acmart}
\usepackage{booktabs} 
\usepackage{graphicx}
\usepackage[export]{adjustbox}
\usepackage{xcolor}
\definecolor{highlight}{RGB}{255, 255, 0}

\usepackage{multirow}
\usepackage{amsmath}
\usepackage{xspace}
\usepackage{xcolor}
\usepackage{dsfont}
\usepackage{rotating}
\usepackage{makecell}

\usepackage{amsthm}
\usepackage{subcaption}
\renewcommand{\epsilon}{\varepsilon}

\usepackage{mathtools}

\newcommand{\R}{\mathbb{R}}

\usepackage{algorithm}
\usepackage{multirow}
\usepackage{algorithmic}

\begin{document}
\title{Comparison of High-Dimensional Bayesian Optimization Algorithms on BBOB}

\author{Maria Laura Santoni}
\orcid{0009-0007-7389-2555}
\affiliation{
  \institution{Sorbonne Universit\'e, CNRS, LIP6}
  \city{Paris}
  \country{France}}
\email{maria-laura.santoni@lip6.fr}

\author{Elena Raponi}
\orcid{0000-0001-6841-7409}
\affiliation{
  \institution{Leiden Institute of Advanced Computer Science, Leiden University}
  \city{Leiden}
  \country{The Netherlands}}
\email{e.raponi@liacs.leidenuniv.nl}

\author{Renato De Leone}
\orcid{0000-0001-9783-608X}
\affiliation{
  \institution{School of Science and Technology, University of Camerino}
  \city{Camerino}
  \country{Italy}}
\email{renato.deleone@unicam.it}

\author{Carola Doerr}
\orcid{0000-0002-4981-3227}
\affiliation{
  \institution{Sorbonne Universit\'e, CNRS, LIP6}
  \city{Paris}
  \country{France}}
\email{carola.doerr@lip6.fr}

\begin{abstract} 

Bayesian Optimization (BO) is a class of surrogate-based black-box optimization heuristics designed to efficiently locate high-quality solutions for problems that are expensive to evaluate, and therefore allow only small evaluation budgets. BO is particularly popular for solving numerical optimization problems in industry, where the evaluation of objective functions often relies on time-consuming simulations or physical experiments. However, many industrial problems depend on a large number of parameters. This poses a challenge for BO algorithms, whose performance is often reported to suffer when the dimension grows beyond 15 decision variables. 
Although many new algorithms have been proposed to address this, it remains unclear which one is best suited for a specific optimization problem.

In this work, we compare five state-of-the-art high-dimensional BO algorithms with vanilla BO, CMA-ES, and random search on the 24 BBOB functions of the COCO environment at increasing dimensionality, ranging from 10 to 60 variables. Our results confirm the superiority of BO over CMA-ES for limited evaluation budgets and suggest that the most promising approach to improve BO is the use of trust regions. However, we also observe significant performance differences for different function landscapes and budget exploitation phases, indicating improvement potential, e.g., through hybridization of algorithmic components. 
\end{abstract}




\maketitle

\sloppy
%%%%%%%%%%%%%%%%%%%%%%
\section{Introduction}
\label{sec:intro}

In countless areas of science, engineering, and beyond, researchers and developers deal with free parameters that can be tuned to achieve a specific goal. With the increase in computing power and simulation methods, optimization techniques have become always more popular in industry. They replace the manual trial-and-error method for parameter tuning, which is highly dependent on prior knowledge and expertise about the particular research area and subject, and which requires many trials before a satisfactory parameter configuration is found, without any guarantee of its optimality. 
In many application areas such as machine learning, neural network design, robotics, aerospace, and mechanical design, optimization deals with black-box problems, i.e., problems where the structure of the objective function, its derivatives, and/or the constraints is unknown, unexploitable, or non-existent.
This is the case, for example, in design and prototyping processes, where engineers simulate the performance of new products with increasingly accurate numerical models and simulations. This makes evaluating a particular parameter setting extremely costly, reducing the affordable budget for simulations and making it impossible to fully explore the design space.
When limited evaluation budgets are available in optimization, it is a common practice to rely on surrogate models, i.e., approximations that mimic the behavior of the expensive and unknown objective function to be optimized while being computationally cheaper to evaluate.

One of the most commonly used surrogate-based optimization methods is Bayesian Optimization (BO)~\cite{mockus2012bayesian, garnett_bayesoptbook_2023}. Through careful modeling -- based on Gaussian process regression (GPR) -- and intelligent search for candidate solutions -- through the optimization of an acquisition function -- BO algorithms can deliver impressive optimization performance even with small evaluation budgets~\cite{garnett_bayesoptbook_2023}.
The potential of BO has been demonstrated in hundreds of studies across a wide range of domains, such as chemistry and material science~\cite{hernandez-lobato_parallel_2017,griffiths_constrained_2020,kotthoff_bayesian_2021}, biology, engineering and design~\cite{lam_advances_2018,raponi_kriging-assisted_2019,raponi_methodology_2021}, robotics~\cite{martinez-cantin_bayesian_2009,calandra_bayesian_2016,junge_improving_2020}, algorithm configuration~\cite{hutter_sequential_2011-1}, hyperparameter tuning~\cite{klein_fast_2017-1,nguyen_bayesian_2019,turner_bayesian_2021-2}, and automated machine learning~\cite{malkomes_bayesian_2016,jin_auto-keras_2019}. 
However, when the dimensionality of the problem exceeds 15 variables, the performance of BO deteriorates due to the so-called \textit{curse of dimensionality}~\cite{bellman_dynamic_1966}. Scaling BO to high-dimensional spaces is challenging, because its statistical and computational complexity increases with dimension: the number of points queried to satisfactorily cover the search space increases exponentially with the dimension, and optimizing the acquisition function requires more and more computational power, being a non-convex optimization problem on the same design space itself. 

\textbf{Related work: } In recent years, characterized by increasingly complex systems and large and high-dimensional data, great efforts have been made to extend BO to higher dimensions. Various strategies have been proposed. According to~\cite{binoisSurveyHighdimensionalGaussian2021}, these are mainly based on (but not limited to) one of the following methods: Variable selection, additive models, linear and nonlinear embeddings, and trust regions. All of these strategies have advantages and disadvantages (see Figure~\ref{HBDC5}), and it is not clear which one is best for which optimization scenario. In Section~\ref{sec:HDBO} we discuss these categories further and give some examples. However, for a more comprehensive overview, we refer the reader to available surveys on this topic \cite{malu_bayesian_2021,binoisSurveyHighdimensionalGaussian2021}, even though they are purely informative and lack an experimental study comparing the methods.
On the other hand, comparative studies have been conducted to present new algorithms such as SMAC~\cite{hutter_evaluation_2013}, REMBO~\cite{wang_bayesian_2016}, TuRBO~\cite{eriksson2019scalable}, SAASBO~\cite{eriksson2021high}, etc., but they are either outdated or limited to showing the potential of the proposed algorithm rather than presenting a comprehensive and unbiased performance comparison.
Our work aims at filling this gap.

\textbf{Disclaimer:} In line with~\cite{binoisSurveyHighdimensionalGaussian2021} we refer to the setting with dimension 60 as \textit{high-dimensional}, even if BO-approaches for problems with several thousands of variables have been studied~\cite{wang_bayesian_2016}.

\textbf{Our contribution: } In this article we present the results of a benchmarking study that follows the standardized, well-established guidelines for unbiased performance comparison in numerical black-box optimization. With the goal to obtain a first overview over which high-dimensional BO (HDBO) approaches to favor for which problem characteristics, we benchmark BO variants that are specifically designed for high-dimensional, low-budget optimization problems. We also include in our comparison three standard baselines: vanilla BO~\cite{head2017scikit}, the Covariance Matrix Adaptation Evolution Strategy (CMA-ES)~\cite{hansen2001completely}, and random search~\cite{harris2020array}. To obtain interpretable results, we focus on the 24 (noiseless) functions of the Black-Box Optimization Benchmarking (BBOB) suite from the COCO environment~\cite{hansen2021coco} and compare algorithm performance, in terms of both convergence speed and CPU time, for what we consider small evaluation budgets; motivated by our target applications, we set the budget to $10D+50$.
It is important to note that our study aims to fairly compare diverse HDBO algorithms by reporting metrics like model fitting time and acquisition function optimization time, utilizing open-source benchmark functions. This provides a leveled ground for algorithm performance analysis and for understanding how it is influenced by the inner mechanisms of algorithm classes.
Key findings from our experiments are that (1) Vanilla BO performs better than CMA-ES for small dimensions and low budget, (2) many of the algorithms that aim to scale BO to high-dimensional spaces outperform vanilla BO and CMA-ES, with the convergence gap widening as the dimension increases, and (3) among the compared algorithms, the BO variant using trust regions performs particularly well on a large number of function, dimension, and budget combinations.

\textbf{Reproducibility:} Our code for reproducing the experiments is available
on GitHub, in the public repository \verb|IOH-HDBO-Comparison|.\footnote{~\url{https://github.com/MariaLauraSantoni/IOH-Profiler-HDBO-Comparison}} 
The project data is also available for interactive analysis and visualization on the IOHanalyzer platform~\cite{IOHprofiler} as `HDBO' dataset. 

Code and full data are also stored in Zenodo under the name of \verb|Comparison of High-Dimensional Bayesian| \verb|Optimization Algorithms on BBOB|.\footnote{~\url{https://zenodo.org/doi/10.5281/zenodo.8099720}}
The collected data also includes results for the \textbf{Heteroscedastic and Evolutionary Bayesian Optimisation (HEBO) solver~\cite{cowenrivers2022hebo}}. This algorithm does not easily fit into one of the five categories proposed by~\cite{binoisSurveyHighdimensionalGaussian2021}, as it does not rely on search space reduction techniques. Instead, it employs nonlinear input and output warping to enhance the quality of the GP surrogates and multi-objective acquisition functions. In our experiments on the 24 BBOB functions, HEBO demonstrates promising performance. Nevertheless, it is excluded from our comparison due to a notable number of unsuccessful runs in the highest dimensions. The data collection also comprises results for \textbf{Ensemble Bayesian Optimization (EBO)}~\cite{wang2018batched}, another HDBO method tested within the `additive models' category. However, this algorithm performed worse than \textbf{Random Decomposition Upper-Confidence Bound (RDUCB)}~\cite{pmlr-v202-ziomek23a}; its results are therefore not shown in this paper. Further details can be found in A.3.

\textbf{Outline of the paper:} Section~\ref{sec:BO} introduces BO. Approaches to make BO efficient in high dimensions are discussed in Section~\ref{sec:HDBO}.
Following this, Section~\ref{sec:setup} presents an overview of the experimental setup employed for our comparative study. 
Results are presented and discussed in Sections~\ref{sec:results} and~\ref{sec:furtherdiscussion}. Finally, Section~\ref{sec:conclusion} summarizes the conclusions drawn from our study and outlines the subsequent important steps for further exploration of the current challenges.
Additional details regarding the individual algorithms and an extensive analysis of the obtained results are provided in the appendix.

%%%%%%%%%%%%%%%%%%%%%%

\section{Bayesian Optimization}
\label{sec:BO}
Bayesian Optimization (BO)~\cite{mockus2012bayesian,frazier2018tutorial,garnett_bayesoptbook_2023} is one of the most widely used surrogate-based optimization approaches. It is a global optimization strategy based on Bayesian inference, tailored to solve optimization problems with a small number of function evaluations.

BO involves two main components: a method for statistical inference, usually a GPR model, and an acquisition function that determines the optimal locations for sampling new points.

The procedure starts with a prior probability distribution for the objective function $f$, usually a Gaussian process prior, and a budget $T$ for the maximum number of function evaluations. Next, the objective function $f$ is evaluated on $n_0$ points that are distributed within the design space according to a Design of Experiments (DoE) scheme~\cite{forrester_engineering_2008}. This process generates an initial sample set $(X,y)=\{(x^1,y^1),(x^2,y^2),...,(x^{n_0},y^{n_0})\}$.
As long as the total budget is not exhausted, a GPR is used to construct a posterior Bayesian probability distribution describing potential values for $y=f(x)$ at a candidate point $x$ in the design space, together with the uncertainty in the prediction. We determine the next point to evaluate by maximizing an acquisition function (e.g., expected improvement, probability of improvement, upper confidence bound, entropy search~\cite{sobesterEngineeringDesignSurrogate2008a}) that indicates how much the evaluation of the proposed point contributes to the optimization goal, i.e., how much potential this point has to improve on the best solution found up to this iteration. 

Depending on the acquisition function definition, the search strategy can show a more exploitative or more explorative attitude, favoring the investigation of areas of the search domain with low predicted mean or high predicted uncertainty, respectively.

We then observe the objective function at that point, i.e., we evaluate $y^n = f(x^n)$, add the new information $(x^n, y^n)$ to the sample set, and update the GPR prediction to obtain a new posterior distribution for the objective function. This process is iterated until a stopping criterion (e.g., exhausted evaluation budget) is met.
Algorithm~\ref{alg:Bayesopt} provides a high-level overview of BO.
%%%%%%%%%%%%%%%%%%%%%%%%%%%%%%%%%%%%%%%%
\begin{algorithm}[t]
\caption{Basic pseudocode for BO}\label{alg:Bayesopt}
\begin{algorithmic}[1] 
\REQUIRE Objective function $f:S \rightarrow \R$, 
total budget $T$, number of initial points $n_0$, acquisition function $\alpha$.
\STATE Place a Gaussian process prior on $f$;

\STATE Create a data set of $n_{0}$ points, $X=\{x^1,x^2,...,x^{n_0}\}$ according to a space-filling DoE scheme;  
\STATE Evaluate $f$ on the $X$, let $y^i=f(x^i)$, and let $y=\{y^1,y^2,...,y^{n_0}\}$; 

\STATE Set $n = n_{0}$;
\WHILE{$n \leq T$}
\STATE Increment $n$ by 1;
\STATE Train the GPR model on the sample set $(X,y)$ to get the posterior probability distribution on $f$;

\STATE Find the next point $x^{n}$ to query by maximizing the acquisition function $\alpha$ using the current posterior distribution: $x^{n} = \operatorname*{argmax}\{\alpha{(x)}|\ x \in S \}$;
\STATE Evaluate the function on the new point and let $y^{n} = f(x^{n})$;
\STATE Increase the data set by updating $(X,y) = (X,y) \cup (x^{n},y^{n})$;
\STATE Update the current-best solution $x^* = \operatorname*{argmin} \left \{ f(x^{i})| 1 \leq i \leq n \right \}$;
\ENDWHILE
\STATE Return $x^*$.
\end{algorithmic}
\end{algorithm}

\section{High-Dimensional Bayesian Optimization}
\label{sec:HDBO}

Several independent studies report that \emph{vanilla BO}, i.e., the standard version of the algorithm, works well up to around 15 decision variables~\cite{zhigljavsky2012theory,zhigljavsky2021bayesian,binoisSurveyHighdimensionalGaussian2021}. For larger dimensions, it becomes inefficient compared to other black-box optimization solvers.

The problem of high dimensionality in BO has been addressed using several strategies that, according 
to~\cite{binoisSurveyHighdimensionalGaussian2021}, can be grouped into five broad categories: 
\begin{itemize}
    \item variable selection~\cite{salem2019sequential,marrel2008efficient,eriksson2021high},
    \item additive models~\cite{durrande2011additive,delbridge2020randomly, wang2018batched, pmlr-v202-ziomek23a},
    \item linear embeddings~\cite{letham2020re,chen2020semi,raponiHighDimensionalBayesian2020a, letham2020reexamining},
    \item nonlinear embeddings~\cite{wycoff2019sequential,gaudrie2020modeling,kPCABO},
    \item trust regions~\cite{eriksson2019scalable, diouane2023trego, regis2016trust}.
\end{itemize}
Researchers do not agree on which approach is best. There are pros and cons to each of them, and their performance is often influenced by the available evaluation budget and the problem structure. For example, the first two categories assume an underlying structure on the objective function, e.g., intrinsic lower dimensionality or additive structures (a decomposition of the objective function into a sum of lower-dimensional components).

In this article, we compare algorithms representing the five different approaches mentioned above. In the selection of these algorithms, preference was given to the most recent ones and those that provide a Python implementation.

For the variable selection approach, we focus on \textbf{Sparse Axis Aligned Subspace Bayesian Optimization (SAASBO)}~\cite{eriksson2021high}. This approach introduces a new surrogate model for high-dimensional BO based on the assumption that the coordinates of $ x$ in the design space $S$ have a hierarchy of relevance.
According to~\cite{eriksson2021high} this approach has several key advantages. First, it preserves the structure of the input space and can therefore exploit it. Second, it is adaptive and shows low sensitivity to its hyperparameters. Third, it can naturally accommodate both input and output constraints, unlike methods based on random projections for which input constraints are a particular challenge.

Among the algorithms that use additive models, we analyze \textbf{Random Decomposition Upper-Confidence Bound (RDUCB)}~\cite{pmlr-v202-ziomek23a}. The main idea of RDUCB is to learn the decomposition of the black-box function by using a data-independent decomposition rule, a randomized tree sampling strategy. An additive acquisition function is used together with an upper confidence bound adjustment in conjunction with a message-passing optimizer. This optimizer is designed to better exploit additive acquisition structures. The algorithm successfully addresses the problem of data-driven strategies and prevents them from being easily misled into a local decomposition that does not apply globally to the entire search space.

Among the linear and nonlinear embeddings approaches, we focus on \textbf{PCA-assisted Bayesian Optimization (PCA-BO)}~\cite{raponiHighDimensionalBayesian2020a} and \textbf{Kernel PCA-assisted Bayesian Optimization (KPCA-BO)}~\cite{kPCABO}, respectively. 

KPCA-BO is an extended version of PCA-BO, which uses kernel methods to first map points to a reproducing kernel Hilbert space (RKHS)~\cite{berlinet2011reproducing} using an implicit nonlinear feature mapping. Then, PCA in the RKHS is used to learn a linear transformation from all evaluated points during the run and select dimensions in the transformed space according to the variability of the evaluated points. The main advantage of KPCA-BO over PCA-BO is the nonlinearity of the submanifold in the search space, which makes it more likely to catch multiple basins of attraction simultaneously.

For trust-region-based approaches, we consider \textbf{Trust Region Bayesian Optimization (TuRBO)}~\cite{eriksson2019scalable}. TuRBO can be defined as a local strategy for BO. It introduces the use of trust regions in BO, making it a technique for global optimization that uses a collection of simultaneous local optimization runs with independent probabilistic models. Thompson sampling~\cite{hernandez2016distributed} is used to find new points to evaluate within a single trust region and across the set of trust regions simultaneously. 
The main advantage of this approach is that each local surrogate model has the typical advantages of Bayesian modeling, i.e., robustness to noisy observations and uncertainty estimates. Moreover, the local surrogates allow heterogeneous modeling of the objective function and do not suffer from overexploration. This local approach is complemented by a global bandit strategy that distributes samples across these confidence regions, implicitly balancing exploration and exploitation.

\begin{figure*}[t] \center
    \includegraphics[width=\textwidth]{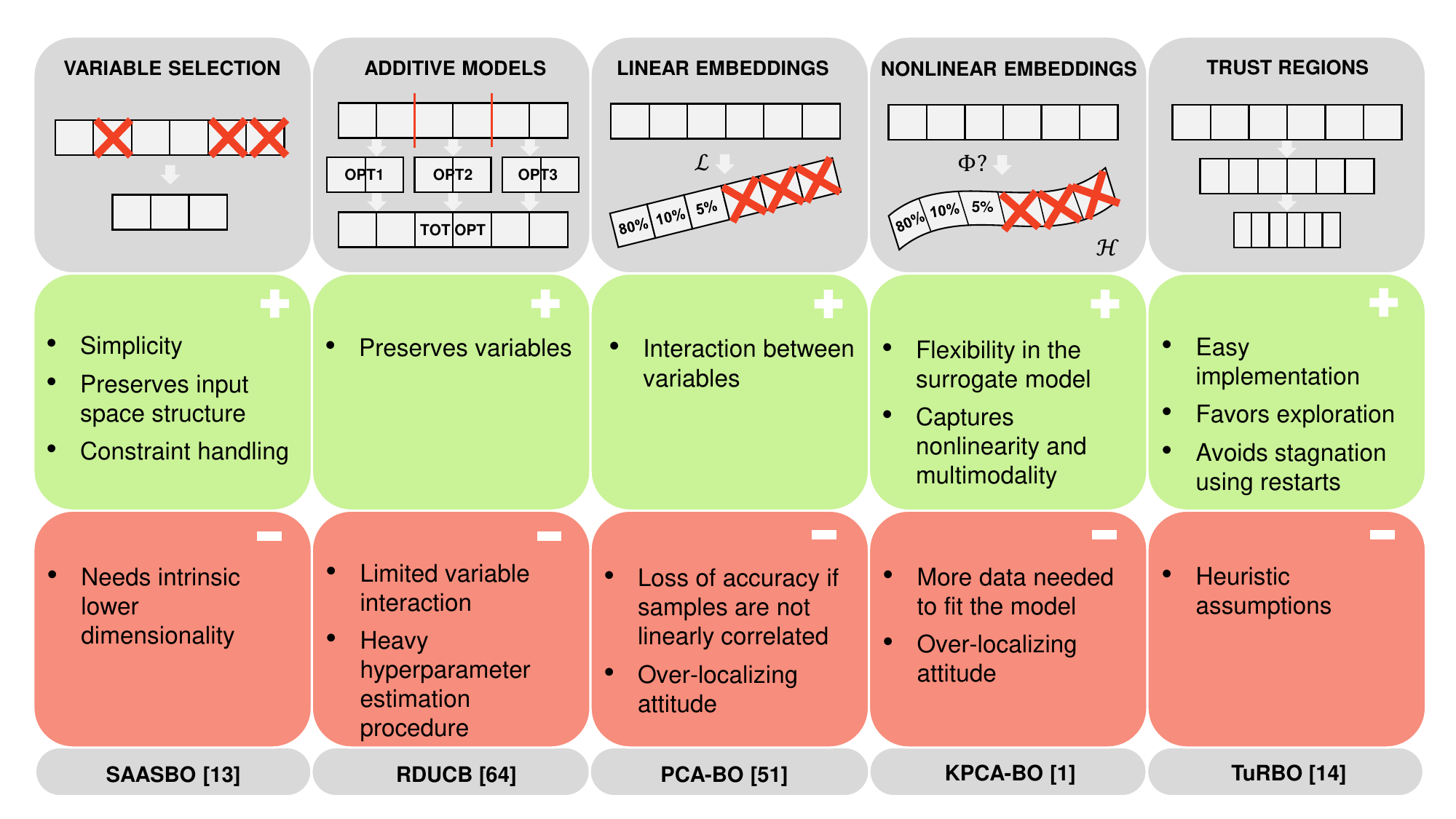}
      \caption{Illustration of the five different categories of BO-based algorithms for high-dimensional problems. The advantages and disadvantages of each category are listed, along with the algorithm chosen for our comparison.}
    \label{HBDC5}
\end{figure*}

Each of the five HDBO classes has advantages and disadvantages, and the same applies to the algorithms that belong to them. 
Figure~\ref{HBDC5} summarizes the idea behind each approach and its pros and cons.
To date, there is no consensus on which method is preferable under which circumstances (problem landscape, evaluation budget, and dimensionality).

%%%%%%%%%%%%%%%%%%%%%%

\section{Experimental Setup}
\label{sec:setup}
We compare the algorithms on the 24 noiseless functions of COCO's BBOB suite~\cite{hansen2021coco}. We access these functions via IOHprofiler~\cite{doerr2018iohprofiler}. After initial exploration of the data with IOHanalyzer, we use Python post-processing libraries to generate the plots visualizing our results in this paper. 

The BBOB functions are divided into five groups: separable functions (f1-f5), functions with low or moderate conditioning (f6-f9), functions with high conditioning and unimodal structure (f10-f14), multimodal functions with appropriate global structure (f15-f19), and multimodal functions with weak global structure (f20-f24)~\cite{bbobfunctions}. In our experience, the problems belonging to the last two groups are most representative of real-world problems~\cite{10.1145/3512290.3528712}, as they have a more complex landscape characterized by high nonlinearity and roughness, with multiple peaks or valleys. For each function, we consider dimensions 10, 20, 40, and 60. For SAASBO we did not run complete experiments for dimensions 20, 40, and 60 due to time and memory constraints. In this case, we preferred to perform experiments on functions belonging to the last two groups (f15-f24). 

Thus, we conducted 10 independent runs of each algorithm for three different instances (instance ID 0-2) of the 24 BBOB functions, with the only exception of SAASBO. For this algorithm, we have:
\begin{itemize}
    \item data for functions f1-f24 at dimension 10,
    \item data for functions f15-f24 at dimensions 20 and 40,
    \item no data for dimension 60 (infeasible amount of computational time and memory required).
\end{itemize}

We compare the HDBO algorithms with a vanilla BO from~\cite{head2017scikit}, a default CMA-ES~\cite{hansen2019cma}, and random search~\cite{harris2020array}.

For each run, the total evaluation budget is set to $10D+50$ function evaluations. For BO-based algorithms, the initial DoE size is set to $D$. 
By default for the BBOB suite, the domain is set to $[-5, 5]^D$.
For each algorithm described in Section~\ref{sec:HDBO}, we use default settings for their hyperparameters.
We evaluate algorithm performance in terms of (1) loss (defined as the target precision, i.e., the absolute difference between the best-so-far value and the value of the global minimum) and (2) CPU time. We run our experiments on the LRZ Linux cluster,\footnote{\url{https://doku.lrz.de/coolmuc-2-11484376.html}} with a 64-bit x86 CPU architecture. The cluster is equipped with Intel Xeon CPU E5-2690 v3 processors running at 2.60 GHz. It consists of 812 nodes, with each node having 28 cores and 64 GB of memory. We use source code as provided by the authors of the respective algorithms: we did not optimize code for CPU efficiency or other criteria. Although we are aware that GPU acceleration is available for certain Python package implementations, we deliberately opted against utilizing it to ensure a fair algorithm runtime comparison. This decision was also driven by our primary focus of investigating BO-based algorithms under constrained evaluation budgets -- a circumstance frequently encountered in real-world applications, where high computational time is usually required for each function evaluation. Consequently, enhancing the computational efficiency of the algorithm, achievable through CUDA and GPU acceleration, was deemed secondary.

The implementation of vanilla BO is taken from the Python module \verb|scikit-optimize|,\footnote{~\url{https://scikit-optimize.github.io/stable/auto_examples/bayesian-optimization.html}} choosing Expected Improvement (EI) as the acquisition function and a noise level equal to $0.01$.
The implementation for CMA-ES~\cite{hansen2019cma} is the one in the pycma package, 
available from the GitHub repository \verb|pycma|.\footnote{~\url{https://github.com/CMA-ES/pycma}}
The implementation for the random search uses the Python module \verb|numpy|\footnote{~\url{https://numpy.org/doc/stable/reference/random/generated/numpy.random.uniform.html}} and in particular the method \textit{random.uniform} that draws samples from a uniform distribution over the half-open interval that includes the lower bound, but excludes upper bound of the search space.
The code for SAASBO is taken from the GitHub repository \verb|saasbo|.\footnote{~\url{https://github.com/martinjankowiak/saasbo}} 

For RDUCB, we use the implementation available at the GitHub repository \verb|RDUCB|;\footnote{~\url{https://github.com/huawei-noah/HEBO/tree/master/RDUCB}} it is based on the code for HEBO~\cite{cowenrivers2022hebo}, another HDBO algorithm mentioned in the introduction.

The code for PCA-BO and KPCA-BO is taken from the GitHub repository \verb|Bayesian-Optimization|.\footnote{~\url{https://github.com/wangronin/Bayesian-Optimization/tree/KPCA-BO}}
The TuRBO code is taken from the GitHub repository \verb|TuRBO|\footnote{~\url{https://github.com/uber-research/TuRBO}}. In our experiments, two different implementations of TuRBO are compared: TuRBO1 and TuRBOm. They differ in the number of trust regions used by the algorithm: $tr=1$ and $tr=\lfloor D/5 \rfloor$, respectively, where $tr$ denotes the number of trust regions and $\lfloor \cdot \rfloor$ denotes the floor function. 
For more details on the individual algorithms and the values used for the hyperparameters, please refer to the appendix. 
The code to run the experiments is a modular framework compatible with IOHprofiler. It is available on GitHub in the repository \verb|IOH-HDBO-Comparison|.\footnote{~\url{https://github.com/MariaLauraSantoni/IOH-Profiler-HDBO-Comparison}} 

We use Python post-processing libraries to present our results and the statistical Wilcoxon signed-rank test to confirm what can be seen through direct inspection of the plots.

To better assess the performance of our optimization algorithms, we also present empirical cumulative distribution function (ECDF) curves in Appendix \ref{sec:ecdf}.

%%%%%%%%%%%%%%%%%%%%%%
\section{Results}
\label{sec:results}
\subsection{Dimension D = 10}
\label{sec:10D}
\subsubsection{Solution Quality}
 \begin{figure}[ht]
    \center
    \includegraphics[width=0.95\columnwidth]{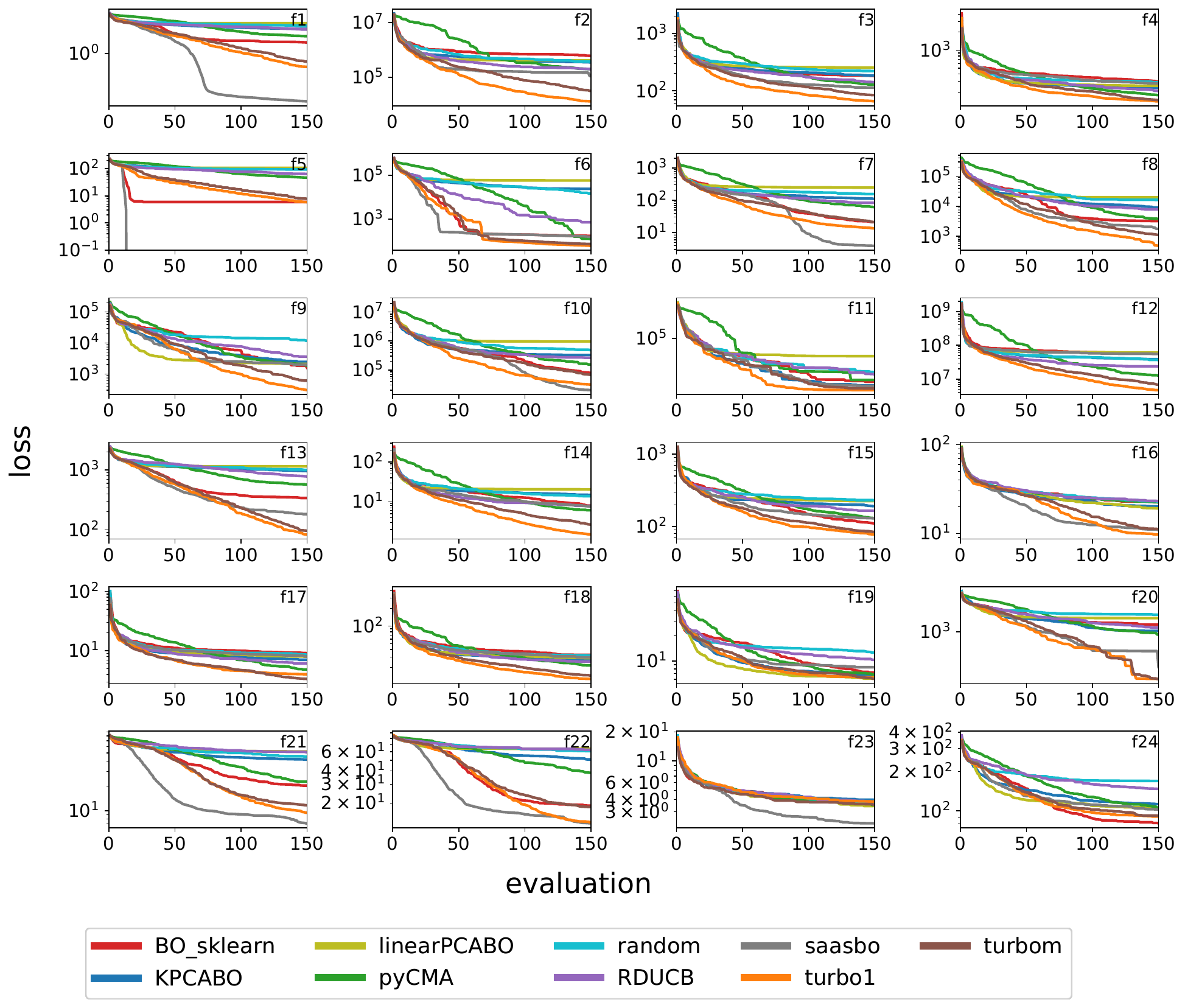}
      \caption{The best-so-far target gap for dimension 10.}
    \label{10D}
\end{figure}
Figure~\ref{10D} compares the convergence behavior of all algorithms on all BBOB functions from f1 to f24 in dimension 10.

In general, random search performed poorly compared to the other algorithms, with only a few exceptions: PCA-BO and KPCA-BO are often comparable or even worse than random search (f1, f3, f4, f6-f8, f10-f12), the same is true for RDUCB (f21, f22) and a few times for vanilla BO, especially for large budgets (f2, f4, f12, f16, f17). This becomes less and less visible with increasing dimensionality, since the purpose of these algorithms is to perform well in high-dimensions.
For a small evaluation budget, vanilla BO always performs better than CMA-ES, as we can see in particular on f2-f4, f12, and f19. However, at the end of the budget, CMA-ES outperforms or is comparable to vanilla BO in many cases.
Overall, we see a good performance of vanilla BO, which is due to the still low dimensionality. BO is always among the best or at least comparable to the other algorithms, with a few exceptions. It reaches a particularly good performance on f5, f6, f22, and f24. 

In Figure~\ref{10D}, algorithm performance is not stable across functions. However, SAASBO and TuRBO tend to predominate. Specifically, TuRBO finds excellent loss values on f2, f3, f13, f14, f16, f18, f20, and f22 (here together with SAASBO).  Stagnation at a very low budget is a common behavior of PCA-BO, KPCA-BO (f1, f5, f6, and f20-f22) and RDUCB (f1, f13, f16, f21, f22).
If we focus on f5 (linear slope), we can observe the extremely good performance of BO and SAASBO, which are able to immediately find the global optimum due to the unimodal and monotonic landscape of this function (this observation holds for all tested dimensions). Finally, we note that sometimes random search outperforms some of the HDBO algorithms (PCA-BO, KPCA-BO, and RDUCB). However, this behavior is not consistent across problems and vanishes in higher dimensionalities, for which HDBO algorithms are specifically designed.
The above observations are confirmed by the ECDF curves in the appendix.

\subsubsection{CPU time}
 \begin{figure}[t] \center \includegraphics[trim=0cm .5cm 0cm 0.3cm, clip, width=\columnwidth]{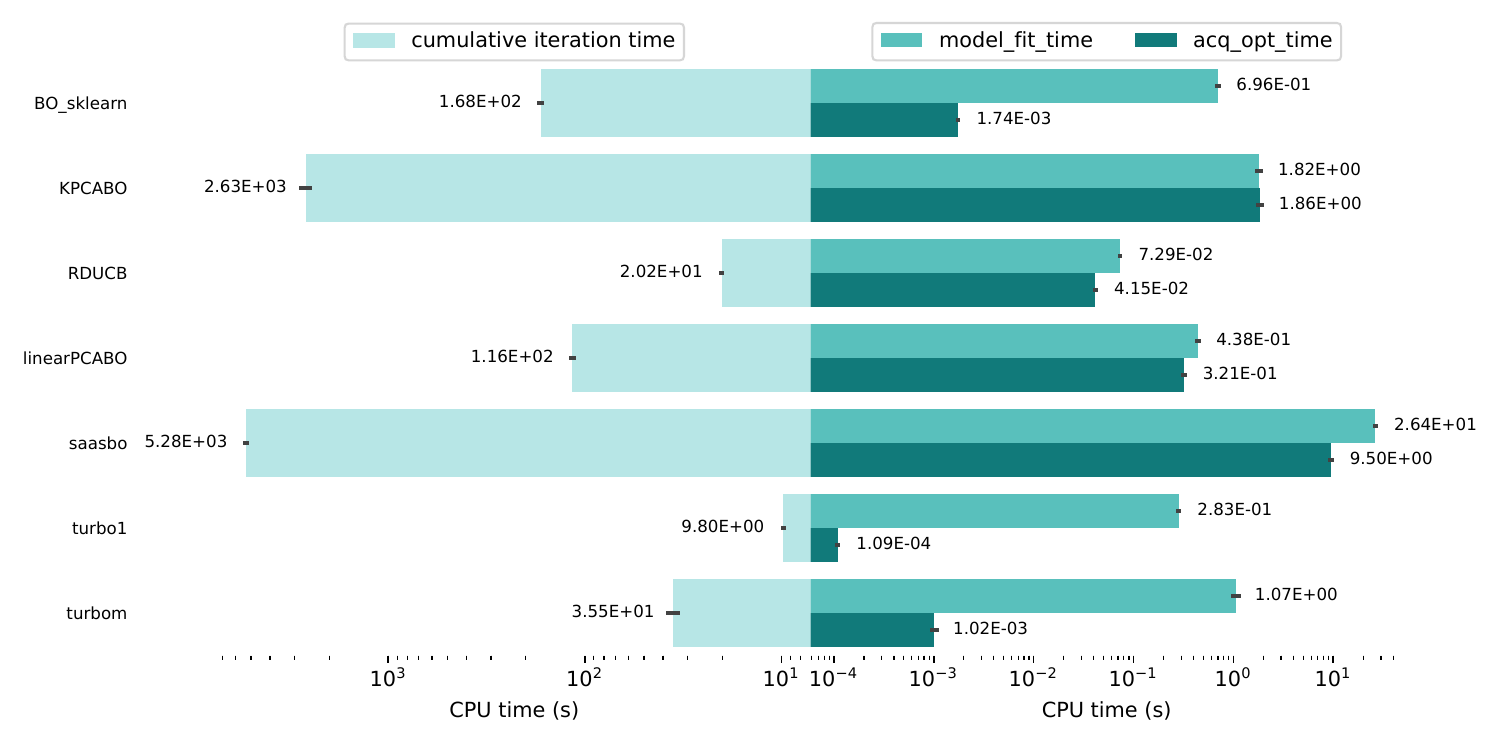} \caption{CPU time in seconds (logarithmic scale) for the entire run (left) and model fitting and acquisition function optimization (right) in dimension 10. Values for the total CPU time are averaged across all 24 BBOB functions. Model fitting time and acquisition function optimization time are first averaged over all iterations of one run, and then across the 24 BBOB functions. The black line in each bar represents the bootstrap confidence interval.} \label{10Dwhole}
\end{figure}

In Figure~\ref{10Dwhole}, we compare the CPU time in seconds (logarithmic scale) to complete the entire run and the CPU time taken by the different algorithms to fit the GPR model and optimize the acquisition function. We also show the bootstrap confidence interval by the black line in each bar~\cite{diciccio1996bootstrap}.
We can clearly see that SAASBO is the most expensive strategy in terms of total CPU time with an average of 5\,284.51 seconds.  
TuRBO1 is the fastest, followed by RDUCB, TuRBOm, and PCA-BO (close to vanilla BO). 
According to the right side of the plot in Figure~\ref{10Dwhole}, SAASBO is also the one that takes more time to complete both steps, fitting the model and optimizing the acquisition function. For almost all algorithms, the CPU time for optimizing the acquisition function is shorter than that for building the model. For PCA-BO and KPCA-BO, they are comparable. From the analyses of convergence and CPU time at 10D, TuRBO1 and TuRBOm seem to perform best.

\subsection{Dimension D = 20}
\label{sec:20D}
\subsubsection{Solution Quality}
 \begin{figure}[h!] \center
    \includegraphics[width=\columnwidth]{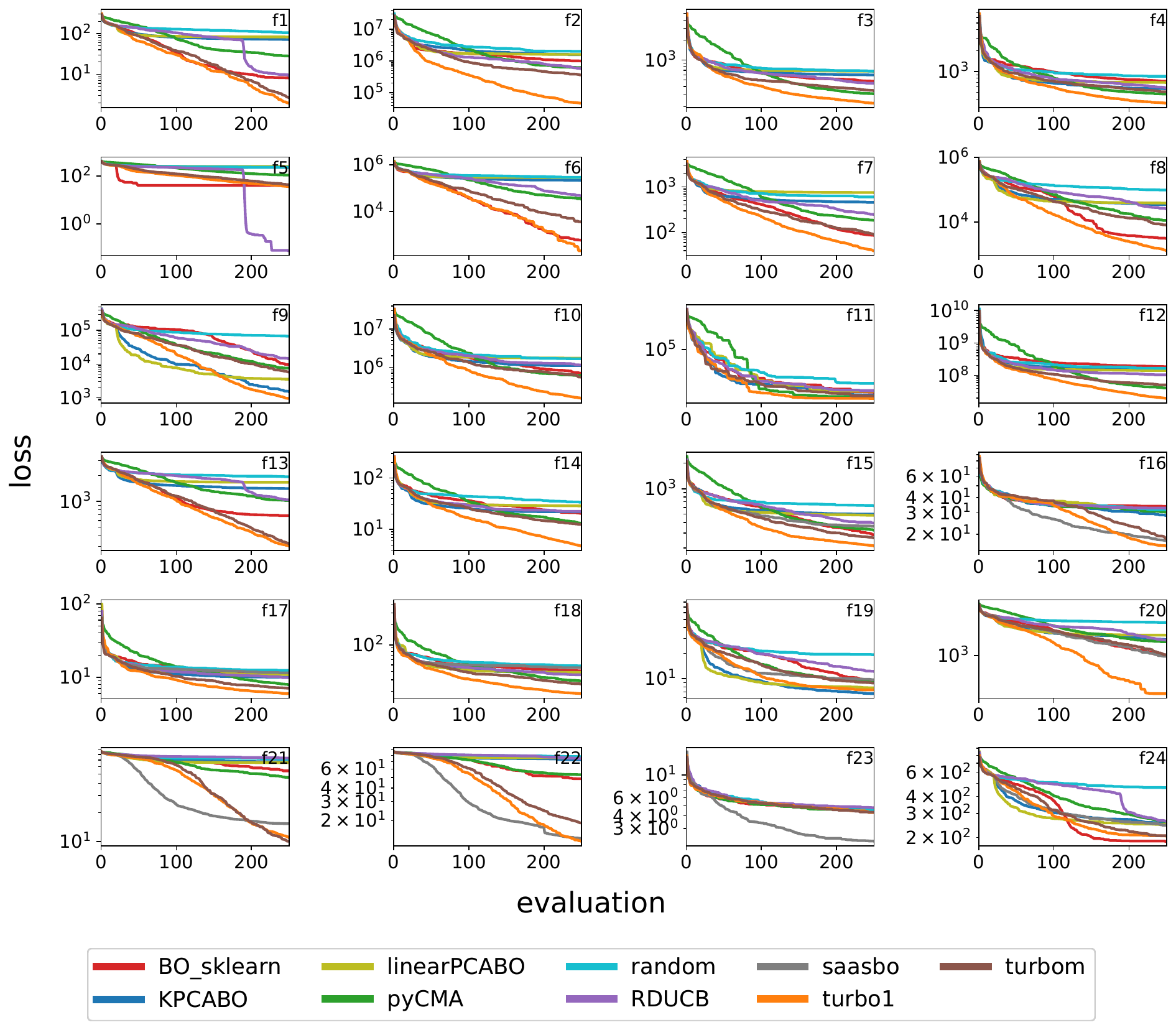}
      \caption{The best-so-far target gap for dimension 20.}
    \label{20D}
\end{figure}
Figure~\ref{20D} compares the convergence behavior of the algorithms on the 20-dimensional problems. We recall that for SAASBO we only have results for f15-f24, due to its high running time. 
With few exceptions, random search performs worse than all tested algorithms. Only for f21 and f22 its performance is comparable to RDUCB, the latter showing stagnant behavior. The overall behavior of the BO algorithms and CMA-ES is very similar to that observed for dimension 10. In particular, we see that vanilla BO always outperforms CMA-ES for small budgets, but CMA-ES catches up for a larger budget of 100 evaluations in many cases (f2-f4, f9, f11, f12, and f14). At a high level, vanilla BO is either better or at least comparable to CMA-ES for the entire budget for all multimodal BBOB functions (f15-f24). We can attribute this to the CMA-ES overlocalizing search attitude.  
It is also interesting to note that at dimension 20, vanilla BO still shows very good performance for some of the functions compared to the BO variants that are specifically designed for high-dimensional problems. In particular, it has one of the best average performances among all algorithms on function f8 and the best one on functions f5 and f24. 
SAASBO and TuRBO1 are the two HDBO variants whose average performance is best among all tested algorithms for the largest number of functions. Their competitive advantage over the other algorithms is clearly visible in the convergence plots in Figure~\ref{20D}, in particular, for f1, f2, f13, and f20-f23.
From this dimension on, we can observe a jump in the convergence behavior of RDUCB that we could not observe for dimension 10 due to the small budget. This jump always occurs around budget 200, often after several evaluations showing a plateau, and it is particularly evident for some functions (f1, f5, f13, f24). For this reason, RDUCB is the best solution for f5 in this dimension. This behavior can also be observed for the plots shown in the original work~\cite{pmlr-v202-ziomek23a}.\footnote{In recent, private communication, the authors attribute this behavior to the \textit{GPy} package used in the original implementation of the algorithm. This package has been replaced by \textit{GPyTorch} in an alternative implementation available at \url{https://github.com/huawei-noah/HEBO/tree/master/MCBO}. However, we have not tested this new implementation and cannot confirm whether it resolves the jumps.}  Nevertheless, RDUCB still shows stagnation on some problems (f16, f21, f22).
The above observations are confirmed by the ECDF curves in the appendix also for dimension 20. Of particular interest is the still good performance of vanilla BO, which continues to show good results, as the dimension is not yet too high for the curse of dimensionality to take effect.  

\subsubsection{CPU time}
 \begin{figure}[t] \center 
 \includegraphics[width=\columnwidth]{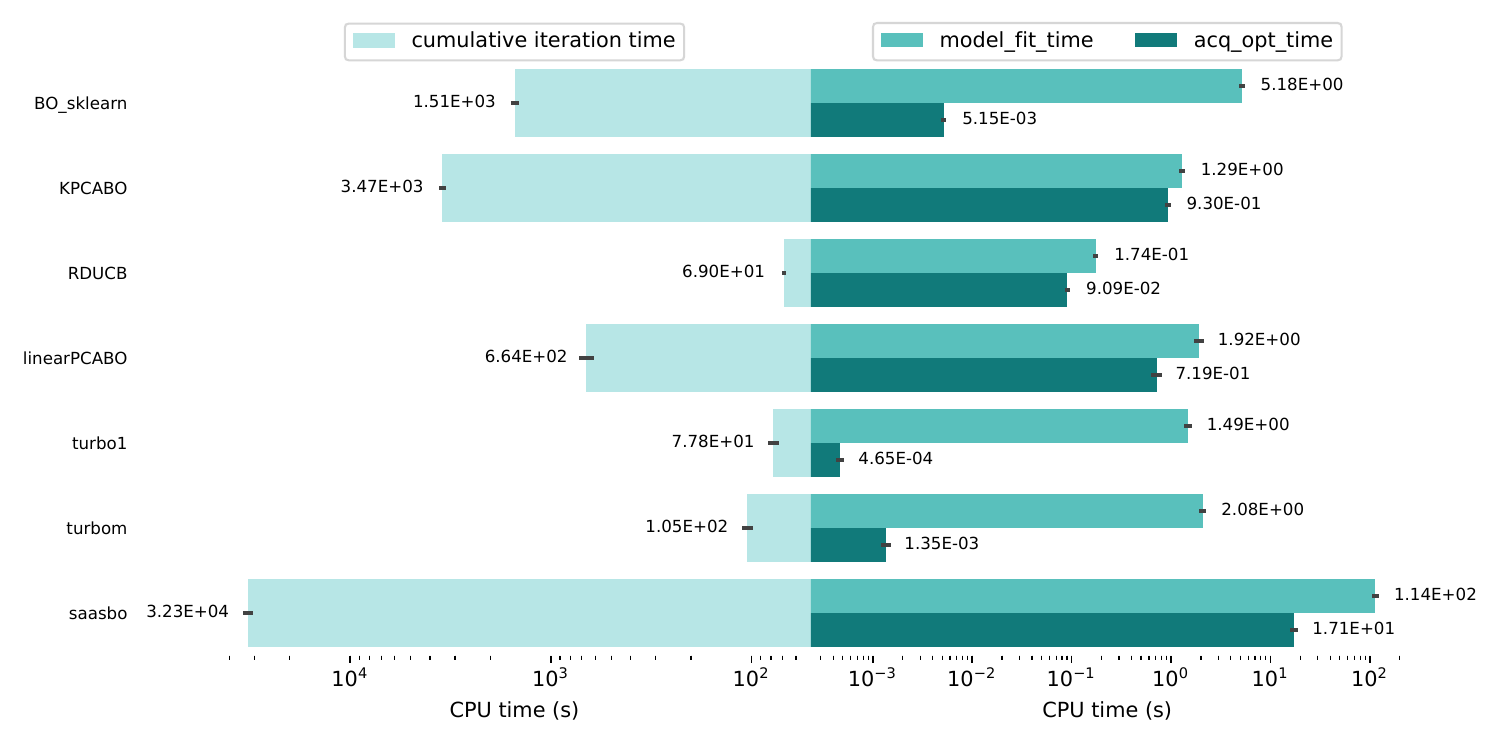} 
 \caption{CPU time in seconds (logarithmic scale) for the entire run (left) and model fitting and acquisition function optimization (right) in dimension 20. Values for the total CPU time are averaged across all 24 BBOB functions. Model fitting time and acquisition function optimization time are first averaged over all iterations of one run, and then across the 24 BBOB functions. The black line in each bar represents the bootstrap confidence interval.}
\label{20Dwhole}
\end{figure}

Figure~\ref{20Dwhole} compares the CPU times for the entire run, for the modeling phase, and for the optimization of the acquisition function for each algorithm tested. As the figure shows, the time increases significantly compared to dimension 10.
For this dimension, the fastest algorithm in terms of cumulative iteration CPU time is RDUCB, followed by TuRBO1 and TuRBOm.
Considering the two different analyses, the experimental results suggest that the best algorithm in dimension 20 is TuRBO1. It seems particularly well suited for f20-f22, which belongs to the category of multimodal functions with weak global structure.

\subsection{Dimension D = 40}
\label{sec:40D}
\subsubsection{Solution Quality}

\begin{figure}[t] \center
    \includegraphics[width=.95\columnwidth]{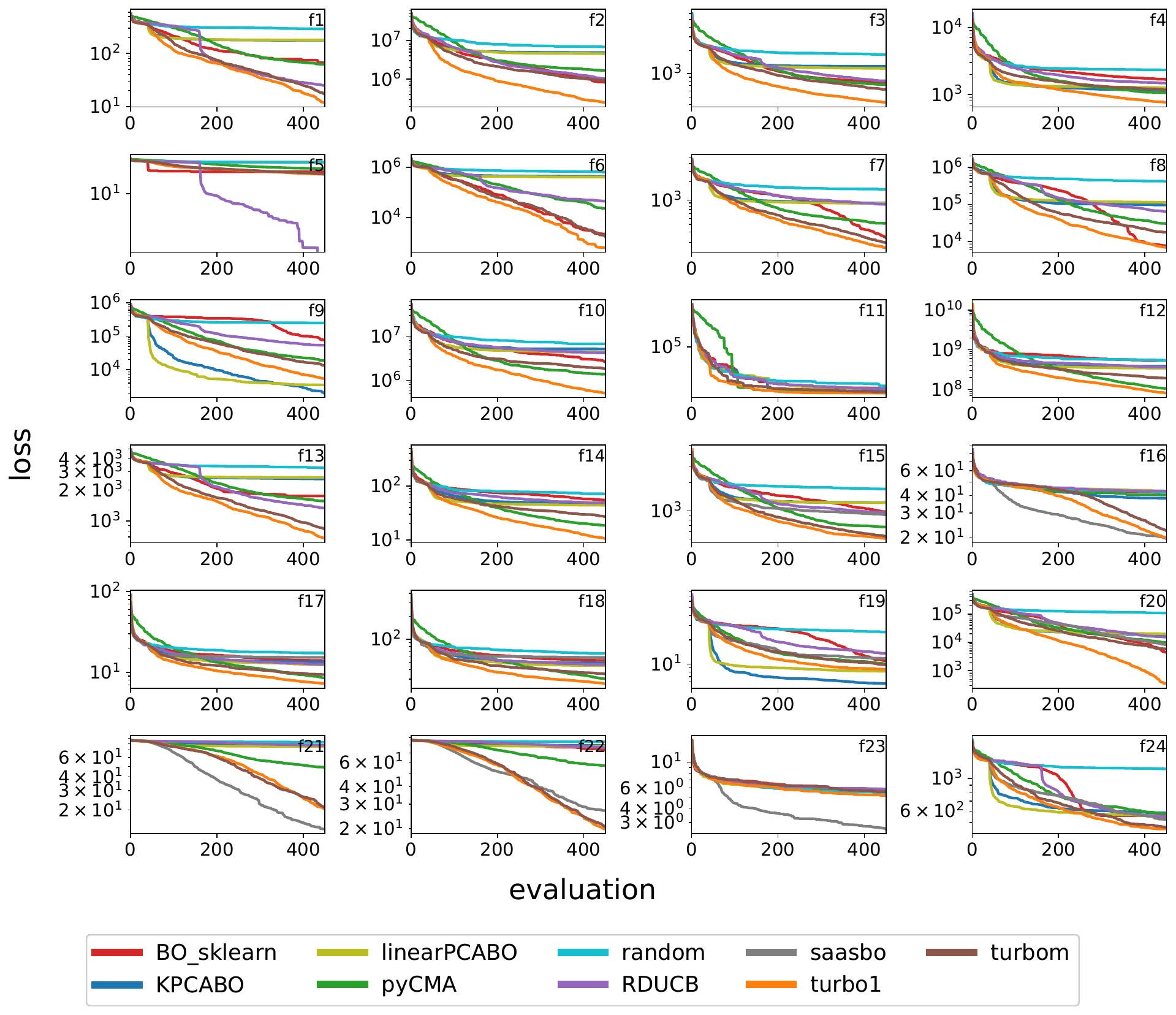}
      \caption{The best-so-far target gap for dimension 40.}
    \label{40D}
\end{figure}

Figure~\ref{40D} shows the convergence plots for dimension 40, where the difference in performance between random search and the other algorithms becomes more visible.
We start seeing that vanilla BO scales badly with increasing dimensionality. For a low budget, of $5D$, i.e., 200 function evaluations, BO is never competitive, with the only exception of f5. The best algorithms remain SAASBO and TuRBO1 for most functions. However, we can also notice that on f9 and f19, the KPCA-BO algorithm, followed by PCA-BO, performs significantly better than all the other methods for the entire evaluation budget. We attribute this to the very good global structure of the function landscapes. In this case, PCA-BO and KPCA-BO are more capable of properly capturing the isocontour of the objective function and it is more likely that basins of attraction are detected.
The ECDF curves in the appendix confirm all previous observations and in particular the effect of the curse of dimensionality for vanilla BO.

\subsubsection{CPU time}
 \begin{figure}[t] \center 
 \includegraphics[trim=0cm 0cm 0cm 0.3cm, clip, width=\columnwidth]{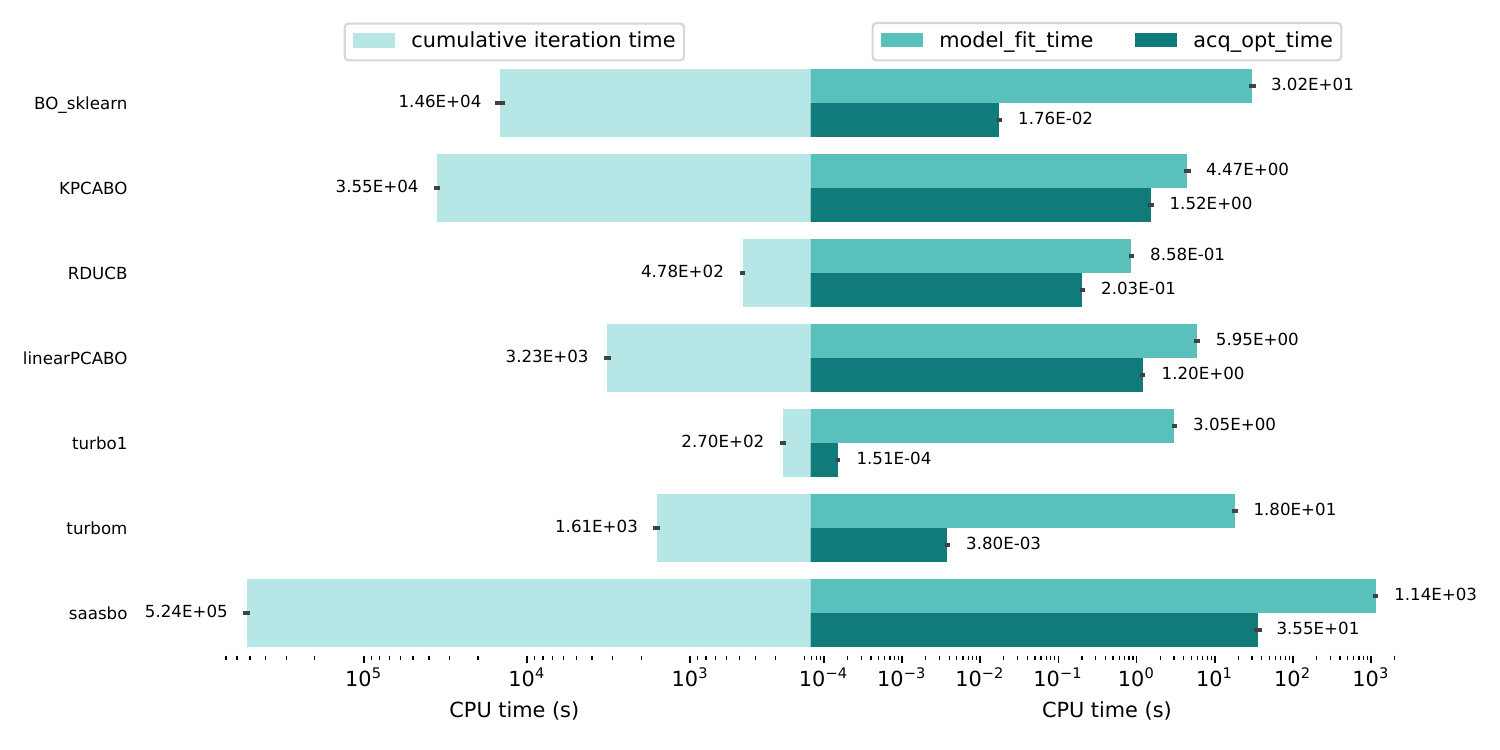} 
 \caption{CPU time in seconds (logarithmic scale) for the entire run (left) and model fitting and acquisition function optimization (right) in dimension 40. Values for the total CPU time are averaged across all 24 BBOB functions.  Model fitting time and acquisition function optimization time are first averaged over all iterations of one run, and then across the 24 BBOB functions. The black line in each bar represents the bootstrap confidence interval.}
 \label{40Dwhole}
\end{figure}

In Figure~\ref{40Dwhole}, the three different CPU times of the algorithms confirm a prohibitive CPU time for SAASBO, while the computational efficiency of TuRBO1 compared to the other algorithms is even more evident than it was for lower dimension. TuRBO is efficient in terms of CPU time because the Thompson Sampling approach substitutes the sub-optimization of a traditional acquisition function. 
RDUCB still maintains a good CPU performance, following directly TuRBO1, and outperforming TuRBOm.

Considering the two different analyses, for dimension 40 we can firmly support the predominance of TuRBO1 over the other algorithms.

\subsection{Dimension D = 60}
\label{sec:60D}
\subsubsection{Solution Quality}
\label{sec:solq60}
 \begin{figure}[ht] \center
    \includegraphics[width=.95\columnwidth]{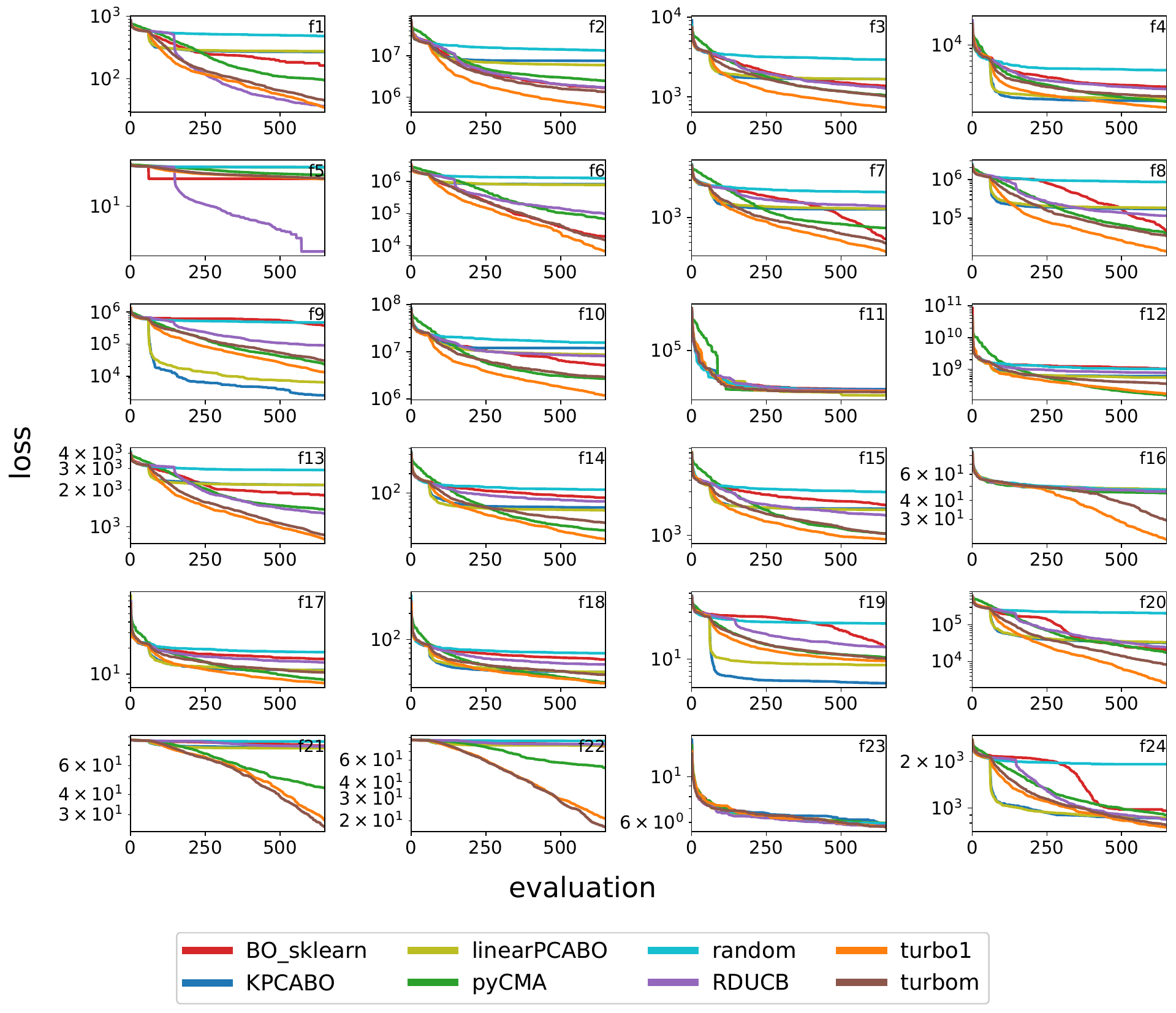}
      \caption{Best-so-far target gap for dimension 60.}
    \label{60D}
\end{figure}

Figure~\ref{60D} compares the convergence behavior of the algorithms for the 60-dimensional BBOB problems. Due to computational constraints, SAASBO is completely missing, as explained in Section~\ref{sec:setup}.
In dimension 60, the inefficiency of random search is obvious, leaving a few exceptions where vanilla BO stagnates and is comparable to it (f9, f12) or where all algorithms are comparable (f11, f16, f21-f23), except for some outliers, such as TuRBO1 and TuRBOm on f16, or also CMA-ES on f21 and f22.
The figure demonstrates how BO suffers from a lower convergence rate at high dimensionality. In all cases except for f5, better convergence capabilities of CMA-ES are evident.
BO suffers from premature stagnation, which is due to its higher computational complexity in dimension 60. 

We can observe the same behavior for some HDBO methods, such as RDUCB and, in some cases, linear and kernel PCA-BO. 
We attribute this to the choice of their hyperparameters as the default ones, which might not be ideal for the function landscapes addressed in this study.
The performance of BO decreases even on f24, where it was the best solver at lower dimensions.
Both versions of TuRBO show very good performance for f1, f13, f21, and f22, and there is a statistically significant difference between them and the other algorithms, which we confirmed by running a Wilcoxon signed-rank test. Moreover, it is worth noting that TuRBO is the only algorithm that continues to improve as the number of evaluations increases, while the other algorithms stagnate easily.
Nonetheless, we can observe interesting performance of PCA-BO and KPCA-BO in some cases. They find excellent loss values on f9, f11, f18-f20, and f24, where they either rank first or show initial speedups that make them good candidates for high-dimension/low-budget optimization problems.

Also in dimension 60, the ECDF curves in the appendix confirm all previous observations. The results highlight the strong impact of high dimensionality on BO, the distinctive and advantageous characteristic of fast convergence for very small evaluation budgets observed for both KPCA-BO and PCA-BO, and the supremacy of TuRBO1 and TuRBOm, which are the only algorithms beating CMA-ES in a fixed-target analysis considering the full evaluation budget.
\subsubsection{CPU time}
 \begin{figure}[ht] \center \includegraphics[width=\linewidth]{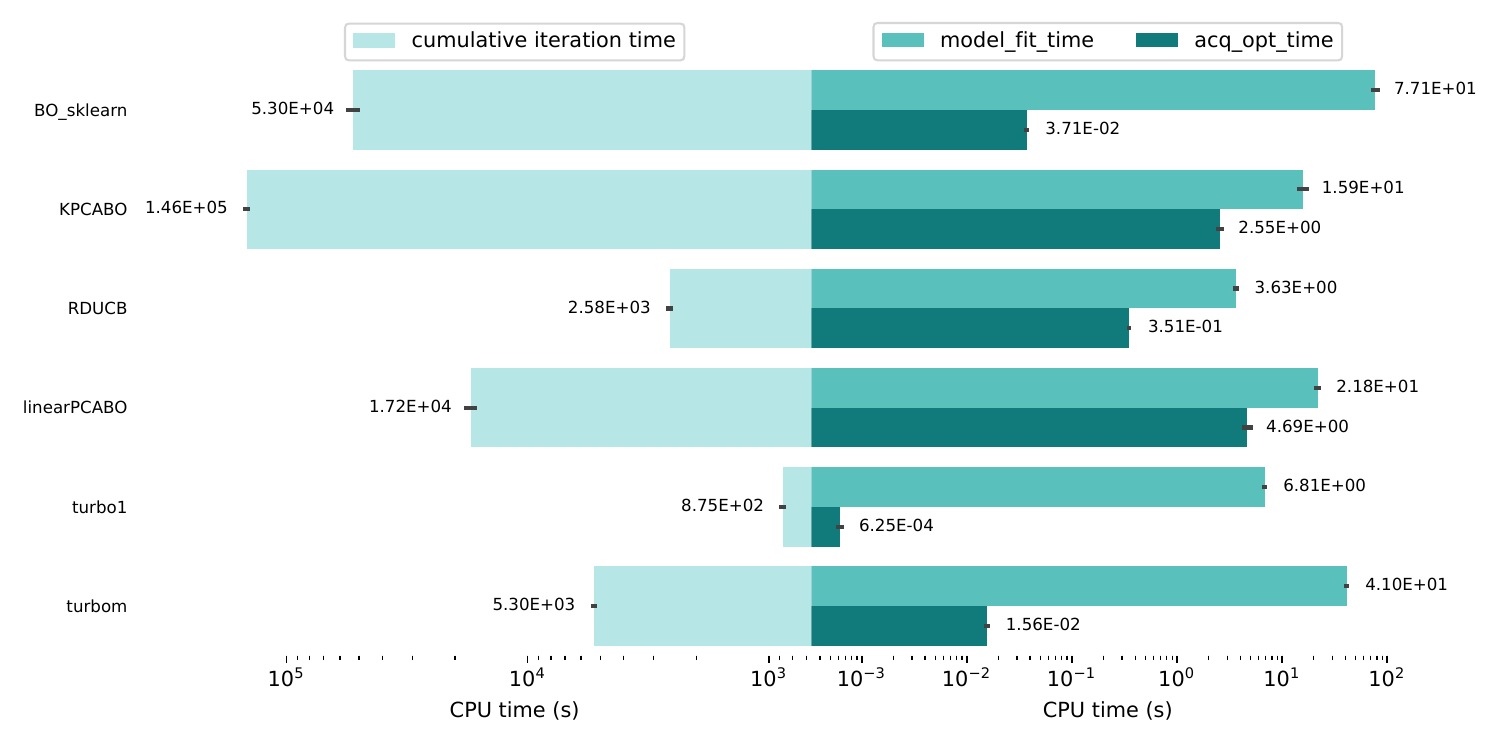} 
\caption{CPU time in seconds (logarithmic scale) for the entire run (left) and model fitting and acquisition function optimization (right) in dimension 60. Values for the total CPU time are averaged across all 24 BBOB functions. Model fitting time and acquisition function optimization time are first averaged over all iterations of one run, and then across the 24 BBOB functions. The black line in each bar represents the bootstrap confidence interval.}
\label{60Dwhole}
\end{figure}
In Figure~\ref{60Dwhole}, on the left, we can see that KPCA-BO and vanilla BO are the algorithms that require the highest CPU time for the run. Again, TuRBO1 is significantly better than the other algorithms and the difference in performance between TuRBO1 and TuRBOm becomes clearer, which suggests the use of just one trust region for high-dimensional problems. 
Once again, we observe RDUCB positioned between TuRBO1 and TuRBOm.
On the right-hand side, Figure~\ref{60Dwhole} confirms that the CPU time for optimizing the acquisition function is generally much shorter than the time required for fitting the model, with almost comparable values for linear and kernel PCA-BO, given that both steps are performed in a mapped space with reduced dimensionality. Finally, we note that the time taken to optimize the acquisition function varies widely between algorithms. This shows that great efforts have been made to improve the efficiency of the infill criterion, which is a main differentiating trait between the algorithms studied.

\section{Further Discussion}
\label{sec:furtherdiscussion}

\begin{figure}[t] \center
    \includegraphics[width=\columnwidth]{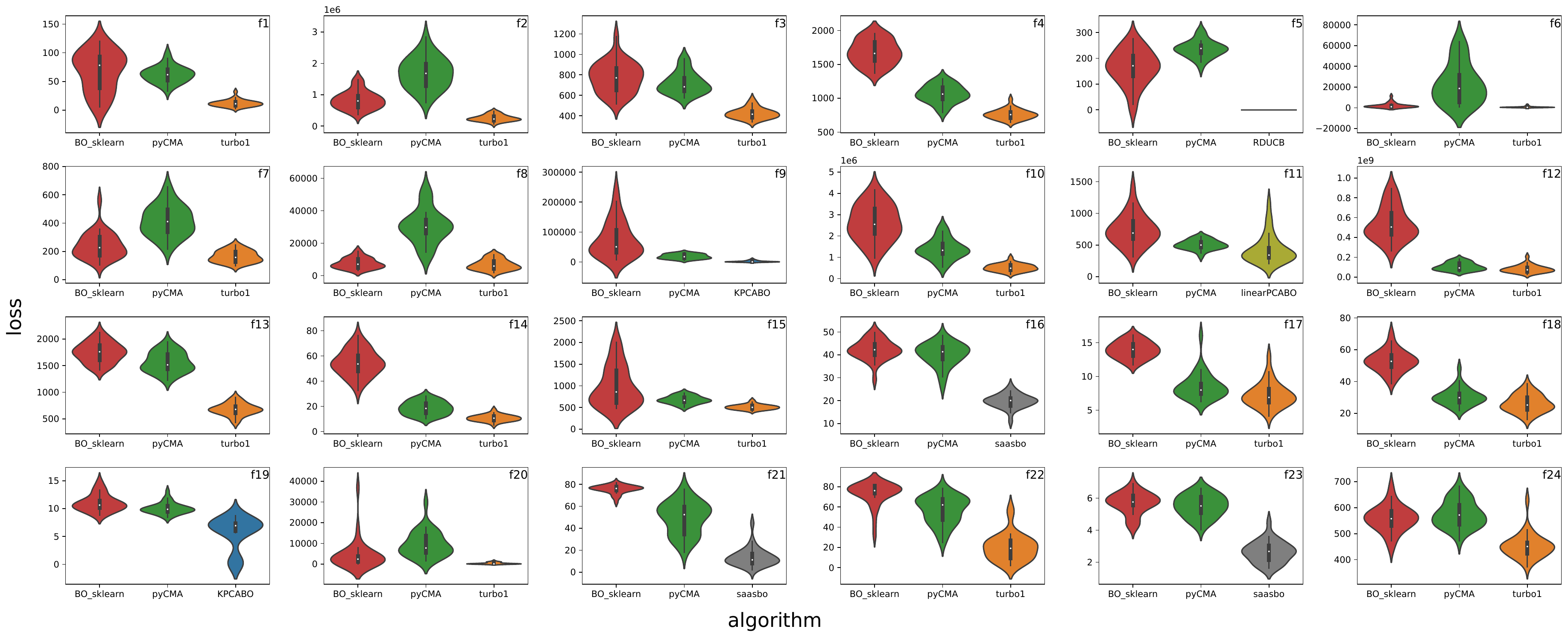}
      \caption{The violin plots show, for dimension 40 and budget 450 (final budget), the distribution of the best-so-far loss values for all functions, obtained by vanilla BO, CMA-ES, and the best among the HDBO algorithms. The plots include a marker for the median of the data and a box indicating the interquartile range.}
    \label{boxplotD40}
\end{figure}

For an in-depth comparison, we also present in Figure~\ref{boxplotD40} violin plots that compare the performance of three candidate algorithms for each BBOB function for dimension 40: vanilla BO, CMA-ES, and the best among the HDBO algorithms for a specific function at the end of the budget (see the appendix for similar violin plots for dimensions 10, 20, and 60). While BO is usually outperformed by CMA-ES at the end of the evaluation budget, the best among the HDBO algorithms always performs better than CMA-ES. 

In general, we can see that TuRBO1 performs the best in dimension 40. However, there are some exceptions: KPCA-BO on f9 and f19, PCA-BO on f11, RDUCB on f5, and SAASBO on f21 and f23. 
The alternative algorithms outperform TuRBO in these functions. This can be attributed to the intrinsic properties of the specific functions, whose landscapes can be better searched through other inner mechanisms than trust regions. For example, some functions have a hierarchical structure in the variables, or they provide a reliable subspace that is well suited for dimensionality reduction by embeddings. In addition, certain algorithms, such as RDUCB or SAASBO (Fig~\ref{10D}), prove to be more effective in dealing with boundary problems. This is particularly evident for f5, where the global optimum lies at the edge of the search space. Therefore, TuRBO demonstrates strong performance across a diverse set of problems in our comparison. However, when specific assumptions about function properties are met, alternative algorithms might prove to be more effective. This emphasizes the importance of choosing the most appropriate algorithm tailored to the specific characteristics of the optimization problem landscape.

 \begin{figure}[t] \center
    \includegraphics[width=0.9\columnwidth]{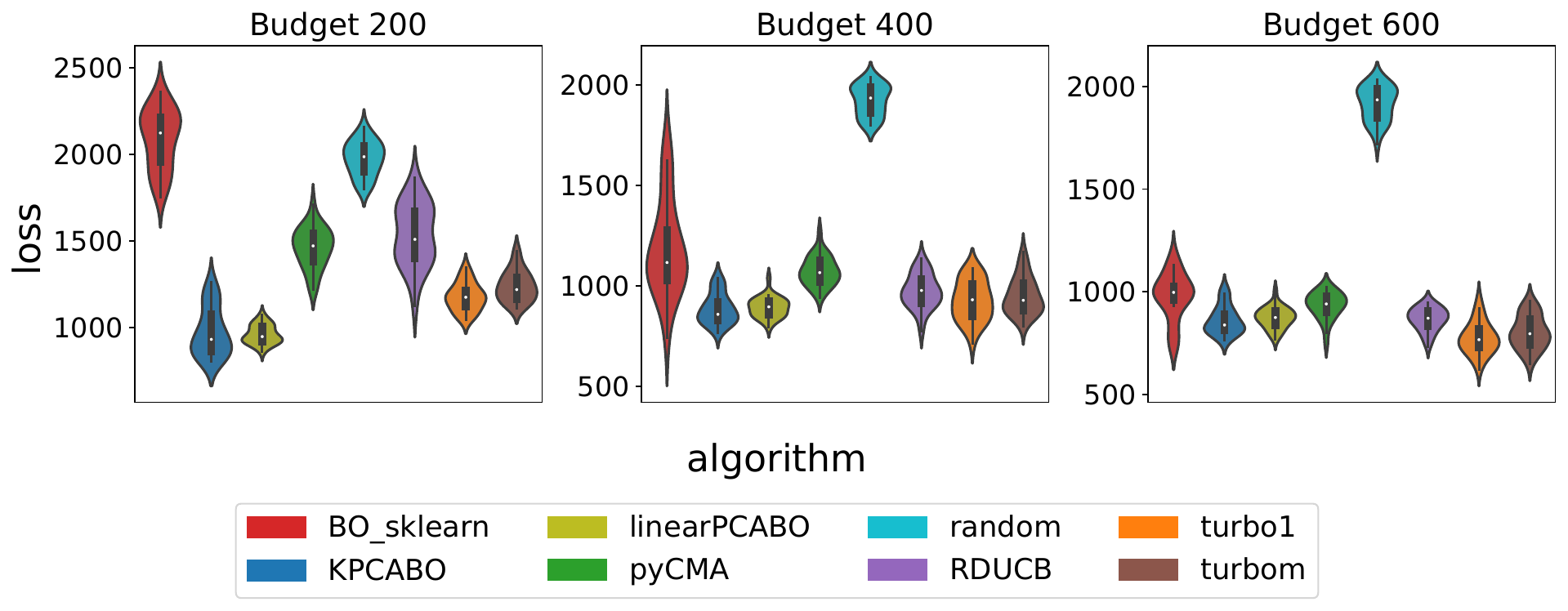}
      \caption{Violin plots showing loss values at three different budgets (200, 400, and 600 function evaluations) for function 24 in dimension 60.}
    \label{f24dim60}
\end{figure}

 Figure~\ref{f24dim60} shows the convergence evolution of the algorithms compared on f24 at dimension 60, by freezing it at three different budgets: 200, 400, and 600 evaluations. The figure gives an idea of the ranking of the algorithms in different phases of the optimization runs and clearly shows in which context one of the algorithms is preferable to the others. For a limited budget (Figure~\ref{f24dim60}, budget 200), PCA-BO and KPCA-BO find the lowest loss values. We attribute this to their better ability to find good solutions in a lower-dimensional manifold of the original search space. However, this often drives the search towards local optima. Already at budget 400 (Figure~\ref{f24dim60}, budget 400), PCA-BO and KPCA-BO become comparable to the other HDBO algorithms and they are significantly outperformed at the end of the run (Figure ~\ref{f24dim60}, budget 600). At this point, TuRBO1 and TuRBOm represent the best choice. 
In fact, TuRBO seems to offer a better balance between exploration and exploitation of the domain. This is due to the use of dynamic trust regions and, in particular, the strategic integration of multiple restarts. The use of random restarts in TuRBO proves particularly beneficial as it allows local optima to be effectively abandoned and exploration to begin anew, increasing its adaptability to a variety of function landscapes. Based on a Wilcoxon signed-rank test, these results are statistically significant. Given these observations, we believe that it would be interesting to explore the possibility of combining the concepts behind PCA-BO and TuRBO, by building local low-dimensional embeddings. In this way, one can benefit from both the flexibility of the trust regions and the lower complexity of the manifolds with reduced dimension. We leave this to our future research.

Finally, although there is no consistent behavior indicating that one algorithm outperforms other algorithms across different BBOB classes, we found that some inner mechanisms are better suited for certain landscape characteristics than others. The main findings are the following: TuRBO performs remarkably well on unimodal landscapes, where its trust regions effectively evolve towards the global basin of attraction (e.g. f2, f3, f13, f14). This trend also extends to multimodal functions with a global unimodal structure and added noise (e.g., f15, f18, f20). However, in the case of multimodal functions with numerous basins of attractions distributed across the design space, a trust region approach, initially guided by a generated sample, runs the risk of neglecting a global perspective and overlooking significant basins of attraction. In such scenarios, SAASBO is a better choice and outperforms other algorithms (e.g. f16, f21-f23). Another interesting case is shown where PCA-BO and KPCA-BO, which use low-dimensional embeddings, show superiority. These algorithms usually perform well in landscapes where the dominant dimension of the basins of attraction easily aligns with the origin of the space (e.g., f9, f19, f24), yet they demonstrate limited generalization capabilities across various functions and instances.

\section{Conclusion and Future Perspectives}
\label{sec:conclusion}
In this work, we have conducted an experimental study comparing the performance of vanilla BO, CMA-ES, random search, and five BO-based algorithms designed to improve BO performance in high-dimensional search spaces. For our tests, we selected one representative for each of the main categories of algorithms for high-dimensional BO: SAASBO for variable selection, RDUCB for additive models, PCA-BO and KPCA-BO for linear and nonlinear embeddings, respectively, and TuRBO for trust regions.
We compared the algorithms on the 24 functions from the open-source BBOB benchmark suite from the COCO benchmarking environment, by running 10 repetitions of each algorithm on 3 different instances of each function, with some exceptions due to time and memory constraints.
With our convergence and CPU time analyses, we provided unbiased and reproducible results for both researchers in the field and practitioners. Researchers can gain insights for algorithm design, while practitioners can leverage our findings for informed algorithm selection in real-world problem-solving. For those looking for a quick overview of HDBO algorithms, our work is a valuable resource that provides a concise and informative introduction to the topic.

Our results confirm a good performance of BO at low dimension (10D), which gradually deteriorates as the dimensionality of the problem increases. Here, CMA-ES performs better, especially for larger budgets. 
However, the average observed performance of CMA-ES is worse than that of the HDBO algorithms.  
Although we observe different performances for different function landscapes and budget utilization phases, TuRBO seems to be the most promising algorithm, both in terms of convergence trend and CPU time. However, PCA-BO and KPCA-BO also show potential for small evaluation budgets, with very fast convergence towards a near-optimal solution.
Therefore, further work is planned to develop a hybrid algorithm that combines PCA-BO and TuRBO. This algorithm could avoid the stagnation of PCA-BO by using restarts and trust regions and still benefit from a linear low-dimensional embedding, resulting in a very competitive algorithm for optimizing expensive black-box functions at high dimensionality.
Indeed, the high potential resulting from the combination of linear embeddings and trust regions has already been demonstrated through the introduction of the Bayesian optimization with adaptively expanding subspaces (BAxUS) algorithm~\cite{papenmeier2022increasing}. This algorithm will serve as a valuable baseline for comparison purposes.

We also found that the poor performance of some algorithms on some functions could be due to poor initialization of their hyperparameters. Hence, we plan to investigate how they could benefit from hyperparameter optimization and repeat our comparative analysis.

%%%%%%%%%%%%%%%%%%%%%%
\begin{acks}
We thank the anonymous reviewers of the manuscript and the reproducibility reviewer for their constructive comments which helped us improve the choice of algorithms to discuss and the ease by which our experiments can be reproduced. We also thank Diederick Vermetten from the University of Leiden for making our data available through the IOHanalyzer website \url{https://iohanalyzer.liacs.nl/}. 

This work was realized with the support of the Sorbonne Center for Artificial Intelligence (SCAI) of Sorbonne University (IDEX SUPER 11-IDEX-0004), of the ANR T-ERC project \emph{VARIATION} (ANR-22-ERCS-0003-01), and of CNRS INS2I project \emph{IOHprofiler}. 
The work leading to this publication was also supported by the PRIME programme of the German
Academic Exchange Service (DAAD) with funds from the German Federal Ministry of Education and
Research (BMBF). 
The contribution of Renato De Leone has been funded by the European Union - NextGenerationEU under the Italian Ministry of University and Research (MUR) National
Innovation Ecosystem grant ECS00000041 - VITALITY - CUP J13C22000430001.
This study was also carried out under the project INdAM – GNCS Project 2023, codice CUP\_E53C22001930001.
 The manuscript reflects only the authors’ views and opinions, neither the European Union nor the European Commission can be considered responsible for them.
\end{acks}
%%%%%%%%%%%%%%%%%%%%%%

\bibliographystyle{ACM-Reference-Format}
\bibliography{arxiv_final} 


\begin{thebibliography}{64}

%%% the Web address.

\ifx \showCODEN    \undefined \def \showCODEN     #1{\unskip}     \fi
\ifx \showDOI      \undefined \def \showDOI       #1{#1}\fi
\ifx \showISBNx    \undefined \def \showISBNx     #1{\unskip}     \fi
\ifx \showISBNxiii \undefined \def \showISBNxiii  #1{\unskip}     \fi
\ifx \showISSN     \undefined \def \showISSN      #1{\unskip}     \fi
\ifx \showLCCN     \undefined \def \showLCCN      #1{\unskip}     \fi
\ifx \shownote     \undefined \def \shownote      #1{#1}          \fi
\ifx \showarticletitle \undefined \def \showarticletitle #1{#1}   \fi
\ifx \showURL      \undefined \def \showURL       {\relax}        \fi
% invisible to TeX
\providecommand\bibfield[2]{#2}
\providecommand\bibinfo[2]{#2}
\providecommand\natexlab[1]{#1}
\providecommand\showeprint[2][]{arXiv:#2}

\bibitem[Antonov et~al\mbox{.}(2022)]%
        {kPCABO}
\bibfield{author}{\bibinfo{person}{Kirill Antonov}, \bibinfo{person}{Elena
  Raponi}, \bibinfo{person}{Hao Wang}, {and} \bibinfo{person}{Carola Doerr}.}
  \bibinfo{year}{2022}\natexlab{}.
\newblock \showarticletitle{High Dimensional Bayesian Optimization with Kernel
  Principal Component Analysis}. In \bibinfo{booktitle}{\emph{Proc. of Parallel
  Problem Solving from Nature (PPSN)}}, Vol.~\bibinfo{volume}{13398}.
  \bibinfo{publisher}{Springer}, \bibinfo{pages}{118--131}.
\newblock
\urldef\tempurl%
\url{https://doi.org/10.1007/978-3-031-14714-2\_9}
\showDOI{\tempurl}


\bibitem[Bellman(1966)]%
        {bellman_dynamic_1966}
\bibfield{author}{\bibinfo{person}{R. Bellman}.}
  \bibinfo{year}{1966}\natexlab{}.
\newblock \showarticletitle{Dynamic Programming}.
\newblock \bibinfo{journal}{\emph{Science (New York, N.Y.)}}
  \bibinfo{volume}{153}, \bibinfo{number}{3731} (\bibinfo{date}{July}
  \bibinfo{year}{1966}), \bibinfo{pages}{34--37}.
\newblock
\showISSN{0036-8075}
\urldef\tempurl%
\url{https://doi.org/10.1126/science.153.3731.34}
\showDOI{\tempurl}


\bibitem[Berlinet and Thomas-Agnan(2011)]%
        {berlinet2011reproducing}
\bibfield{author}{\bibinfo{person}{Alain Berlinet} {and}
  \bibinfo{person}{Christine Thomas-Agnan}.} \bibinfo{year}{2011}\natexlab{}.
\newblock \bibinfo{booktitle}{\emph{Reproducing kernel Hilbert spaces in
  probability and statistics}}.
\newblock \bibinfo{publisher}{Springer Science \& Business Media}.
\newblock


\bibitem[Binois and Wycoff(2022)]%
        {binoisSurveyHighdimensionalGaussian2021}
\bibfield{author}{\bibinfo{person}{Micka{\"{e}}l Binois} {and}
  \bibinfo{person}{Nathan Wycoff}.} \bibinfo{year}{2022}\natexlab{}.
\newblock \showarticletitle{A Survey on High-dimensional Gaussian Process
  Modeling with Application to Bayesian Optimization}.
\newblock \bibinfo{journal}{\emph{{ACM} Trans. Evol. Learn. Optim.}}
  \bibinfo{volume}{2}, \bibinfo{number}{2} (\bibinfo{year}{2022}),
  \bibinfo{pages}{8:1--8:26}.
\newblock
\urldef\tempurl%
\url{https://doi.org/10.1145/3545611}
\showDOI{\tempurl}


\bibitem[Calandra et~al\mbox{.}(2016)]%
        {calandra_bayesian_2016}
\bibfield{author}{\bibinfo{person}{Roberto Calandra},
  \bibinfo{person}{Andr{\'e} Seyfarth}, \bibinfo{person}{Jan Peters}, {and}
  \bibinfo{person}{Marc~Peter Deisenroth}.} \bibinfo{year}{2016}\natexlab{}.
\newblock \showarticletitle{Bayesian Optimization for Learning Gaits under
  Uncertainty}.
\newblock \bibinfo{journal}{\emph{Annals of Mathematics and Artificial
  Intelligence}} \bibinfo{volume}{76}, \bibinfo{number}{1}
  (\bibinfo{date}{Feb.} \bibinfo{year}{2016}), \bibinfo{pages}{5--23}.
\newblock
\showISSN{1573-7470}
\urldef\tempurl%
\url{https://doi.org/10.1007/s10472-015-9463-9}
\showDOI{\tempurl}


\bibitem[Chen et~al\mbox{.}(2020)]%
        {chen2020semi}
\bibfield{author}{\bibinfo{person}{Jingfan Chen}, \bibinfo{person}{Guanghui
  Zhu}, \bibinfo{person}{Rong Gu}, \bibinfo{person}{Chunfeng Yuan}, {and}
  \bibinfo{person}{Yihua Huang}.} \bibinfo{year}{2020}\natexlab{}.
\newblock \showarticletitle{Semi-supervised Embedding Learning for
  High-dimensional Bayesian Optimization}.
\newblock \bibinfo{journal}{\emph{CoRR}}  \bibinfo{volume}{abs/2005.14601}
  (\bibinfo{year}{2020}).
\newblock
\showeprint[arXiv]{2005.14601}
\urldef\tempurl%
\url{https://arxiv.org/abs/2005.14601}
\showURL{%
\tempurl}


\bibitem[Cowen-Rivers et~al\mbox{.}(2022)]%
        {cowenrivers2022hebo}
\bibfield{author}{\bibinfo{person}{Alexander~I. Cowen-Rivers},
  \bibinfo{person}{Wenlong Lyu}, \bibinfo{person}{Rasul Tutunov},
  \bibinfo{person}{Zhi Wang}, \bibinfo{person}{Antoine Grosnit},
  \bibinfo{person}{Ryan~Rhys Griffiths}, \bibinfo{person}{Alexandre~Max
  Maraval}, \bibinfo{person}{Hao Jianye}, \bibinfo{person}{Jun Wang},
  \bibinfo{person}{Jan Peters}, {and} \bibinfo{person}{Haitham~Bou Ammar}.}
  \bibinfo{year}{2022}\natexlab{}.
\newblock \bibinfo{title}{HEBO Pushing The Limits of Sample-Efficient
  Hyperparameter Optimisation}.
\newblock
\newblock
\showeprint[arxiv]{2012.03826}~[cs.LG]


\bibitem[Delbridge et~al\mbox{.}(2020)]%
        {delbridge2020randomly}
\bibfield{author}{\bibinfo{person}{Ian Delbridge}, \bibinfo{person}{David
  Bindel}, {and} \bibinfo{person}{Andrew~Gordon Wilson}.}
  \bibinfo{year}{2020}\natexlab{}.
\newblock \showarticletitle{Randomly projected additive Gaussian processes for
  regression}. In \bibinfo{booktitle}{\emph{International Conference on Machine
  Learning}}. PMLR, \bibinfo{pages}{2453--2463}.
\newblock


\bibitem[DiCiccio and Efron(1996)]%
        {diciccio1996bootstrap}
\bibfield{author}{\bibinfo{person}{Thomas~J DiCiccio} {and}
  \bibinfo{person}{Bradley Efron}.} \bibinfo{year}{1996}\natexlab{}.
\newblock \showarticletitle{Bootstrap confidence intervals}.
\newblock \bibinfo{journal}{\emph{Statistical science}} \bibinfo{volume}{11},
  \bibinfo{number}{3} (\bibinfo{year}{1996}), \bibinfo{pages}{189--228}.
\newblock


\bibitem[Diouane et~al\mbox{.}(2023)]%
        {diouane2023trego}
\bibfield{author}{\bibinfo{person}{Youssef Diouane}, \bibinfo{person}{Victor
  Picheny}, \bibinfo{person}{Rodolophe~Le Riche}, {and}
  \bibinfo{person}{Alexandre Scotto~Di Perrotolo}.}
  \bibinfo{year}{2023}\natexlab{}.
\newblock \showarticletitle{TREGO: a trust-region framework for efficient
  global optimization}.
\newblock \bibinfo{journal}{\emph{Journal of Global Optimization}}
  \bibinfo{volume}{86}, \bibinfo{number}{1} (\bibinfo{year}{2023}),
  \bibinfo{pages}{1--23}.
\newblock


\bibitem[Doerr et~al\mbox{.}(2018)]%
        {doerr2018iohprofiler}
\bibfield{author}{\bibinfo{person}{Carola Doerr}, \bibinfo{person}{Hao Wang},
  \bibinfo{person}{Furong Ye}, \bibinfo{person}{Sander van Rijn}, {and}
  \bibinfo{person}{Thomas B{\"a}ck}.} \bibinfo{year}{2018}\natexlab{}.
\newblock \showarticletitle{{IOHprofiler}: A benchmarking and profiling tool
  for iterative optimization heuristics}.
\newblock \bibinfo{journal}{\emph{arXiv preprint arXiv:1810.05281}}
  (\bibinfo{year}{2018}).
\newblock
\newblock
\shownote{\url{https://iohprofiler.github.io/}}.


\bibitem[Durrande et~al\mbox{.}(2011)]%
        {durrande2011additive}
\bibfield{author}{\bibinfo{person}{Nicolas Durrande}, \bibinfo{person}{David
  Ginsbourger}, {and} \bibinfo{person}{Olivier Roustant}.}
  \bibinfo{year}{2011}\natexlab{}.
\newblock \showarticletitle{Additive kernels for Gaussian process modeling}.
\newblock \bibinfo{journal}{\emph{arXiv preprint arXiv:1103.4023}}
  (\bibinfo{year}{2011}).
\newblock


\bibitem[Eriksson and Jankowiak(2021)]%
        {eriksson2021high}
\bibfield{author}{\bibinfo{person}{David Eriksson} {and}
  \bibinfo{person}{Martin Jankowiak}.} \bibinfo{year}{2021}\natexlab{}.
\newblock \showarticletitle{High-dimensional Bayesian optimization with sparse
  axis-aligned subspaces}. In \bibinfo{booktitle}{\emph{Uncertainty in
  Artificial Intelligence}}. PMLR, \bibinfo{pages}{493--503}.
\newblock


\bibitem[Eriksson et~al\mbox{.}(2019)]%
        {eriksson2019scalable}
\bibfield{author}{\bibinfo{person}{David Eriksson}, \bibinfo{person}{Michael
  Pearce}, \bibinfo{person}{Jacob Gardner}, \bibinfo{person}{Ryan~D Turner},
  {and} \bibinfo{person}{Matthias Poloczek}.} \bibinfo{year}{2019}\natexlab{}.
\newblock \showarticletitle{Scalable global optimization via local {B}ayesian
  {O}ptimization}.
\newblock \bibinfo{journal}{\emph{Advances in Neural Information Processing
  Systems}}  \bibinfo{volume}{32} (\bibinfo{year}{2019}).
\newblock


\bibitem[Forrester et~al\mbox{.}(2008)]%
        {forrester_engineering_2008}
\bibfield{author}{\bibinfo{person}{Alexander I.~J. Forrester},
  \bibinfo{person}{Andr{\'a}s S{\'o}bester}, {and} \bibinfo{person}{Andy~J.
  Keane}.} \bibinfo{year}{2008}\natexlab{}.
\newblock \bibinfo{booktitle}{\emph{Engineering {{Design}} via {{Surrogate
  Modelling}} - {{A Practical Guide}}}}.
\newblock \bibinfo{publisher}{{John Wiley \& Sons Ltd.}}
\newblock
\showISBNx{978-0-470-06068-1}


\bibitem[Frazier(2018)]%
        {frazier2018tutorial}
\bibfield{author}{\bibinfo{person}{Peter~I Frazier}.}
  \bibinfo{year}{2018}\natexlab{}.
\newblock \showarticletitle{A tutorial on Bayesian optimization}.
\newblock \bibinfo{journal}{\emph{arXiv preprint arXiv:1807.02811}}
  (\bibinfo{year}{2018}).
\newblock


\bibitem[Garnett(2023)]%
        {garnett_bayesoptbook_2023}
\bibfield{author}{\bibinfo{person}{Roman Garnett}.}
  \bibinfo{year}{2023}\natexlab{}.
\newblock \bibinfo{booktitle}{\emph{{Bayesian Optimization}}}.
\newblock \bibinfo{publisher}{Cambridge University Press}.
\newblock


\bibitem[Gaudrie et~al\mbox{.}(2020)]%
        {gaudrie2020modeling}
\bibfield{author}{\bibinfo{person}{David Gaudrie}, \bibinfo{person}{Rodolphe
  Le~Riche}, \bibinfo{person}{Victor Picheny}, \bibinfo{person}{Benoit Enaux},
  {and} \bibinfo{person}{Vincent Herbert}.} \bibinfo{year}{2020}\natexlab{}.
\newblock \showarticletitle{Modeling and optimization with Gaussian processes
  in reduced eigenbases}.
\newblock \bibinfo{journal}{\emph{Structural and Multidisciplinary
  Optimization}} \bibinfo{volume}{61}, \bibinfo{number}{6}
  (\bibinfo{year}{2020}), \bibinfo{pages}{2343--2361}.
\newblock


\bibitem[Griffiths and {Hern{\'a}ndez-Lobato}(2020)]%
        {griffiths_constrained_2020}
\bibfield{author}{\bibinfo{person}{Ryan-Rhys Griffiths} {and}
  \bibinfo{person}{Jos{\'e}~Miguel {Hern{\'a}ndez-Lobato}}.}
  \bibinfo{year}{2020}\natexlab{}.
\newblock \showarticletitle{Constrained {{Bayesian}} Optimization for Automatic
  Chemical Design Using Variational Autoencoders}.
\newblock \bibinfo{journal}{\emph{Chemical Science}} \bibinfo{volume}{11},
  \bibinfo{number}{2} (\bibinfo{date}{Jan.} \bibinfo{year}{2020}),
  \bibinfo{pages}{577--586}.
\newblock
\showISSN{2041-6539}
\urldef\tempurl%
\url{https://doi.org/10.1039/C9SC04026A}
\showDOI{\tempurl}


\bibitem[Hansen(2006)]%
        {hansen2006cma}
\bibfield{author}{\bibinfo{person}{Nikolaus Hansen}.}
  \bibinfo{year}{2006}\natexlab{}.
\newblock \showarticletitle{The CMA evolution strategy: a comparing review}.
\newblock \bibinfo{journal}{\emph{Towards a new evolutionary computation}}
  (\bibinfo{year}{2006}), \bibinfo{pages}{75--102}.
\newblock


\bibitem[Hansen et~al\mbox{.}(2019)]%
        {hansen2019cma}
\bibfield{author}{\bibinfo{person}{Nikolaus Hansen}, \bibinfo{person}{Youhei
  Akimoto}, {and} \bibinfo{person}{Petr Baudis}.}
  \bibinfo{year}{2019}\natexlab{}.
\newblock \bibinfo{title}{CMA-ES/pycma on Github. Zenodo, DOI: 10.5281/zenodo.
  2559634.(Feb. 2019)}.
\newblock
\newblock


\bibitem[Hansen et~al\mbox{.}(2021)]%
        {hansen2021coco}
\bibfield{author}{\bibinfo{person}{Nikolaus Hansen}, \bibinfo{person}{Anne
  Auger}, \bibinfo{person}{Raymond Ros}, \bibinfo{person}{Olaf Mersmann},
  \bibinfo{person}{Tea Tusar}, {and} \bibinfo{person}{Dimo Brockhoff}.}
  \bibinfo{year}{2021}\natexlab{}.
\newblock \showarticletitle{{COCO}: A Platform for Comparing Continuous
  Optimizers in a Black-Box Setting}.
\newblock \bibinfo{journal}{\emph{Optimization Methods and Software}}
  \bibinfo{volume}{36} (\bibinfo{year}{2021}), \bibinfo{pages}{114--144}.
\newblock
Issue 1.
\urldef\tempurl%
\url{https://doi.org/10.1080/10556788.2020.1808977}
\showDOI{\tempurl}


\bibitem[Hansen et~al\mbox{.}(2009)]%
        {bbobfunctions}
\bibfield{author}{\bibinfo{person}{Nikolaus Hansen}, \bibinfo{person}{Steffen
  Finck}, \bibinfo{person}{Raymond Ros}, {and} \bibinfo{person}{Anne Auger}.}
  \bibinfo{year}{2009}\natexlab{}.
\newblock \bibinfo{booktitle}{\emph{{Real-Parameter Black-Box Optimization
  Benchmarking 2009: Noiseless Functions Definitions}}}.
\newblock \bibinfo{type}{{T}echnical {R}eport} RR-6829.
  \bibinfo{institution}{{INRIA}}.
\newblock
\urldef\tempurl%
\url{https://hal.inria.fr/inria-00362633/document}
\showURL{%
\tempurl}


\bibitem[Hansen and Kern(2004)]%
        {hansen2004evaluating}
\bibfield{author}{\bibinfo{person}{Nikolaus Hansen} {and}
  \bibinfo{person}{Stefan Kern}.} \bibinfo{year}{2004}\natexlab{}.
\newblock \showarticletitle{Evaluating the CMA evolution strategy on multimodal
  test functions}. In \bibinfo{booktitle}{\emph{International conference on
  parallel problem solving from nature}}. Springer, \bibinfo{pages}{282--291}.
\newblock


\bibitem[Hansen et~al\mbox{.}(2003)]%
        {hansen2003reducing}
\bibfield{author}{\bibinfo{person}{Nikolaus Hansen}, \bibinfo{person}{Sibylle~D
  M{\"u}ller}, {and} \bibinfo{person}{Petros Koumoutsakos}.}
  \bibinfo{year}{2003}\natexlab{}.
\newblock \showarticletitle{Reducing the time complexity of the derandomized
  evolution strategy with covariance matrix adaptation (CMA-ES)}.
\newblock \bibinfo{journal}{\emph{Evolutionary computation}}
  \bibinfo{volume}{11}, \bibinfo{number}{1} (\bibinfo{year}{2003}),
  \bibinfo{pages}{1--18}.
\newblock


\bibitem[Hansen and Ostermeier(2001)]%
        {hansen2001completely}
\bibfield{author}{\bibinfo{person}{Nikolaus Hansen} {and}
  \bibinfo{person}{Andreas Ostermeier}.} \bibinfo{year}{2001}\natexlab{}.
\newblock \showarticletitle{Completely derandomized self-adaptation in
  evolution strategies}.
\newblock \bibinfo{journal}{\emph{Evolutionary computation}}
  \bibinfo{volume}{9}, \bibinfo{number}{2} (\bibinfo{year}{2001}),
  \bibinfo{pages}{159--195}.
\newblock


\bibitem[Harris et~al\mbox{.}(2020)]%
        {harris2020array}
\bibfield{author}{\bibinfo{person}{Charles~R. Harris},
  \bibinfo{person}{K.~Jarrod Millman}, \bibinfo{person}{St{\'{e}}fan~J. van~der
  Walt}, \bibinfo{person}{Ralf Gommers}, \bibinfo{person}{Pauli Virtanen},
  \bibinfo{person}{David Cournapeau}, \bibinfo{person}{Eric Wieser},
  \bibinfo{person}{Julian Taylor}, \bibinfo{person}{Sebastian Berg},
  \bibinfo{person}{Nathaniel~J. Smith}, \bibinfo{person}{Robert Kern},
  \bibinfo{person}{Matti Picus}, \bibinfo{person}{Stephan Hoyer},
  \bibinfo{person}{Marten~H. van Kerkwijk}, \bibinfo{person}{Matthew Brett},
  \bibinfo{person}{Allan Haldane}, \bibinfo{person}{Jaime~Fern{\'{a}}ndez del
  R{\'{i}}o}, \bibinfo{person}{Mark Wiebe}, \bibinfo{person}{Pearu Peterson},
  \bibinfo{person}{Pierre G{\'{e}}rard-Marchant}, \bibinfo{person}{Kevin
  Sheppard}, \bibinfo{person}{Tyler Reddy}, \bibinfo{person}{Warren Weckesser},
  \bibinfo{person}{Hameer Abbasi}, \bibinfo{person}{Christoph Gohlke}, {and}
  \bibinfo{person}{Travis~E. Oliphant}.} \bibinfo{year}{2020}\natexlab{}.
\newblock \showarticletitle{Array programming with {NumPy}}.
\newblock \bibinfo{journal}{\emph{Nature}} \bibinfo{volume}{585},
  \bibinfo{number}{7825} (\bibinfo{date}{Sept.} \bibinfo{year}{2020}),
  \bibinfo{pages}{357--362}.
\newblock
\urldef\tempurl%
\url{https://doi.org/10.1038/s41586-020-2649-2}
\showDOI{\tempurl}


\bibitem[Head et~al\mbox{.}(2017)]%
        {head2017scikit}
\bibfield{author}{\bibinfo{person}{Tim Head}, \bibinfo{person}{M Kumar},
  \bibinfo{person}{H Nahrstaedt}, \bibinfo{person}{G Louppe}, {and}
  \bibinfo{person}{I Shcherbatyi}.} \bibinfo{year}{2017}\natexlab{}.
\newblock \bibinfo{title}{scikit-optimize: Sequential model-based optimization
  in Python}.
\newblock
\newblock


\bibitem[Hern{\'a}ndez-Lobato et~al\mbox{.}(2016)]%
        {hernandez2016distributed}
\bibfield{author}{\bibinfo{person}{Jos{\'e}~Miguel Hern{\'a}ndez-Lobato},
  \bibinfo{person}{Edward Pyzer-Knapp}, \bibinfo{person}{Alan Aspuru-Guzik},
  {and} \bibinfo{person}{Ryan~P Adams}.} \bibinfo{year}{2016}\natexlab{}.
\newblock \showarticletitle{Distributed Thompson Sampling for Large-scale
  Accelerated Exploration of Chemical Space}. In \bibinfo{booktitle}{\emph{NIPS
  Workshop on Bayesian Optimization}}.
\newblock


\bibitem[{Hern{\'a}ndez-Lobato} et~al\mbox{.}(2017)]%
        {hernandez-lobato_parallel_2017}
\bibfield{author}{\bibinfo{person}{Jos{\'e}~Miguel {Hern{\'a}ndez-Lobato}},
  \bibinfo{person}{James Requeima}, \bibinfo{person}{Edward~O. {Pyzer-Knapp}},
  {and} \bibinfo{person}{Al{\'a}n {Aspuru-Guzik}}.}
  \bibinfo{year}{2017}\natexlab{}.
\newblock \showarticletitle{Parallel and {{Distributed Thompson Sampling}} for
  {{Large-scale Accelerated Exploration}} of {{Chemical Space}}}. In
  \bibinfo{booktitle}{\emph{Proceedings of the 34th {{International
  Conference}} on {{Machine Learning}}}}. \bibinfo{publisher}{{PMLR}},
  \bibinfo{pages}{1470--1479}.
\newblock
\showISSN{2640-3498}


\bibitem[Hutter et~al\mbox{.}(2013)]%
        {hutter_evaluation_2013}
\bibfield{author}{\bibinfo{person}{Frank Hutter}, \bibinfo{person}{Holger
  Hoos}, {and} \bibinfo{person}{Kevin {Leyton-Brown}}.}
  \bibinfo{year}{2013}\natexlab{}.
\newblock \showarticletitle{An Evaluation of Sequential Model-Based
  Optimization for Expensive Blackbox Functions}. In
  \bibinfo{booktitle}{\emph{Proc. GECCO (Companion)}}.
  \bibinfo{publisher}{ACM}, \bibinfo{pages}{1209--1216}.
\newblock
\showISBNx{978-1-4503-1964-5}
\urldef\tempurl%
\url{https://doi.org/10.1145/2464576.2501592}
\showDOI{\tempurl}


\bibitem[Hutter et~al\mbox{.}(2011)]%
        {hutter_sequential_2011-1}
\bibfield{author}{\bibinfo{person}{Frank Hutter}, \bibinfo{person}{Holger~H.
  Hoos}, {and} \bibinfo{person}{Kevin {Leyton-Brown}}.}
  \bibinfo{year}{2011}\natexlab{}.
\newblock \showarticletitle{Sequential Model-Based Optimization for General
  Algorithm Configuration}. In \bibinfo{booktitle}{\emph{Proceedings of the 5th
  International Conference on {{Learning}} and {{Intelligent Optimization}}}}
  \emph{(\bibinfo{series}{{{LION}}'05})}.
  \bibinfo{publisher}{{Springer-Verlag}}, \bibinfo{address}{{Berlin,
  Heidelberg}}, \bibinfo{pages}{507--523}.
\newblock
\showISBNx{978-3-642-25565-6}
\urldef\tempurl%
\url{https://doi.org/10.1007/978-3-642-25566-3_40}
\showDOI{\tempurl}


\bibitem[Jin et~al\mbox{.}(2019)]%
        {jin_auto-keras_2019}
\bibfield{author}{\bibinfo{person}{Haifeng Jin}, \bibinfo{person}{Qingquan
  Song}, {and} \bibinfo{person}{Xia Hu}.} \bibinfo{year}{2019}\natexlab{}.
\newblock \bibinfo{title}{Auto-{{Keras}}: {{An Efficient Neural Architecture
  Search System}}}.
\newblock
\newblock
\urldef\tempurl%
\url{https://doi.org/10.48550/arXiv.1806.10282}
\showDOI{\tempurl}
\showeprint[arxiv]{1806.10282}~[cs, stat]


\bibitem[Junge et~al\mbox{.}(2020)]%
        {junge_improving_2020}
\bibfield{author}{\bibinfo{person}{Kai Junge}, \bibinfo{person}{Josie Hughes},
  \bibinfo{person}{Thomas~George Thuruthel}, {and} \bibinfo{person}{Fumiya
  Iida}.} \bibinfo{year}{2020}\natexlab{}.
\newblock \showarticletitle{Improving {{Robotic Cooking Using Batch Bayesian
  Optimization}}}.
\newblock \bibinfo{journal}{\emph{IEEE Robotics and Automation Letters}}
  \bibinfo{volume}{5}, \bibinfo{number}{2} (\bibinfo{date}{April}
  \bibinfo{year}{2020}), \bibinfo{pages}{760--765}.
\newblock
\showISSN{2377-3766, 2377-3774}
\urldef\tempurl%
\url{https://doi.org/10.1109/LRA.2020.2965418}
\showDOI{\tempurl}


\bibitem[Klein et~al\mbox{.}(2017)]%
        {klein_fast_2017-1}
\bibfield{author}{\bibinfo{person}{Aaron Klein}, \bibinfo{person}{Stefan
  Falkner}, \bibinfo{person}{Simon Bartels}, \bibinfo{person}{Philipp Hennig},
  {and} \bibinfo{person}{Frank Hutter}.} \bibinfo{year}{2017}\natexlab{}.
\newblock \bibinfo{title}{Fast {{Bayesian Optimization}} of {{Machine Learning
  Hyperparameters}} on {{Large Datasets}}}.
\newblock
\newblock
\urldef\tempurl%
\url{https://doi.org/10.48550/arXiv.1605.07079}
\showDOI{\tempurl}
\showeprint[arxiv]{1605.07079}~[cs, stat]


\bibitem[Kotthoff et~al\mbox{.}(2021)]%
        {kotthoff_bayesian_2021}
\bibfield{author}{\bibinfo{person}{Lars Kotthoff}, \bibinfo{person}{Hud Wahab},
  {and} \bibinfo{person}{Patrick Johnson}.} \bibinfo{year}{2021}\natexlab{}.
\newblock \showarticletitle{Bayesian {{Optimization}} in {{Materials Science}}:
  {{A Survey}}}.
\newblock \bibinfo{journal}{\emph{arXiv:2108.00002 [cond-mat,
  physics:physics]}} (\bibinfo{date}{July} \bibinfo{year}{2021}).
\newblock
\showeprint[arxiv]{2108.00002}~[cond-mat, physics:physics]


\bibitem[Lam et~al\mbox{.}(2018)]%
        {lam_advances_2018}
\bibfield{author}{\bibinfo{person}{R{\'e}mi Lam}, \bibinfo{person}{Matthias
  Poloczek}, \bibinfo{person}{Peter Frazier}, {and} \bibinfo{person}{Karen~E.
  Willcox}.} \bibinfo{year}{2018}\natexlab{}.
\newblock \showarticletitle{Advances in {{Bayesian Optimization}} with
  {{Applications}} in {{Aerospace Engineering}}}. In
  \bibinfo{booktitle}{\emph{2018 {{AIAA Non-Deterministic Approaches
  Conference}}}}. \bibinfo{publisher}{{American Institute of Aeronautics and
  Astronautics}}, \bibinfo{address}{{Kissimmee, Florida}}.
\newblock
\showISBNx{978-1-62410-529-6}
\urldef\tempurl%
\url{https://doi.org/10.2514/6.2018-1656}
\showDOI{\tempurl}


\bibitem[Letham et~al\mbox{.}(2020a)]%
        {letham2020re}
\bibfield{author}{\bibinfo{person}{Ben Letham}, \bibinfo{person}{Roberto
  Calandra}, \bibinfo{person}{Akshara Rai}, {and} \bibinfo{person}{Eytan
  Bakshy}.} \bibinfo{year}{2020}\natexlab{a}.
\newblock \showarticletitle{Re-examining linear embeddings for high-dimensional
  bayesian optimization}.
\newblock \bibinfo{journal}{\emph{Advances in neural information processing
  systems}}  \bibinfo{volume}{33} (\bibinfo{year}{2020}),
  \bibinfo{pages}{1546--1558}.
\newblock


\bibitem[Letham et~al\mbox{.}(2020b)]%
        {letham2020reexamining}
\bibfield{author}{\bibinfo{person}{Benjamin Letham}, \bibinfo{person}{Roberto
  Calandra}, \bibinfo{person}{Akshara Rai}, {and} \bibinfo{person}{Eytan
  Bakshy}.} \bibinfo{year}{2020}\natexlab{b}.
\newblock \bibinfo{title}{Re-Examining Linear Embeddings for High-Dimensional
  Bayesian Optimization}.
\newblock
\newblock
\showeprint[arxiv]{2001.11659}~[stat.ML]


\bibitem[Long et~al\mbox{.}(2022)]%
        {10.1145/3512290.3528712}
\bibfield{author}{\bibinfo{person}{Fu~Xing Long}, \bibinfo{person}{Bas van
  Stein}, \bibinfo{person}{Moritz Frenzel}, \bibinfo{person}{Peter Krause},
  \bibinfo{person}{Markus Gitterle}, {and} \bibinfo{person}{Thomas B\"{a}ck}.}
  \bibinfo{year}{2022}\natexlab{}.
\newblock \showarticletitle{Learning the Characteristics of Engineering
  Optimization Problems with Applications in Automotive Crash}. In
  \bibinfo{booktitle}{\emph{Proceedings of the Genetic and Evolutionary
  Computation Conference}} (Boston, Massachusetts)
  \emph{(\bibinfo{series}{GECCO '22})}. \bibinfo{publisher}{Association for
  Computing Machinery}, \bibinfo{address}{New York, NY, USA},
  \bibinfo{pages}{1227–1236}.
\newblock
\showISBNx{9781450392372}
\urldef\tempurl%
\url{https://doi.org/10.1145/3512290.3528712}
\showDOI{\tempurl}


\bibitem[Malkomes et~al\mbox{.}(2016)]%
        {malkomes_bayesian_2016}
\bibfield{author}{\bibinfo{person}{Gustavo Malkomes}, \bibinfo{person}{Charles
  Schaff}, {and} \bibinfo{person}{Roman Garnett}.}
  \bibinfo{year}{2016}\natexlab{}.
\newblock \showarticletitle{Bayesian Optimization for Automated Model
  Selection}. In \bibinfo{booktitle}{\emph{Advances in {{Neural Information
  Processing Systems}}}}, Vol.~\bibinfo{volume}{29}.
  \bibinfo{publisher}{{Curran Associates, Inc.}}
\newblock


\bibitem[Malu et~al\mbox{.}(2021)]%
        {malu_bayesian_2021}
\bibfield{author}{\bibinfo{person}{Mohit Malu}, \bibinfo{person}{Gautam
  Dasarathy}, {and} \bibinfo{person}{Andreas Spanias}.}
  \bibinfo{year}{2021}\natexlab{}.
\newblock \showarticletitle{Bayesian {{Optimization}} in {{High-Dimensional
  Spaces}}: {{A Brief Survey}}}. In \bibinfo{booktitle}{\emph{{{International
  Conference}} on {{Information}}, {{Intelligence}}, {{Systems}} \&
  {{Applications}} ({{IISA}})}}. \bibinfo{publisher}{{IEEE}},
  \bibinfo{pages}{1--8}.
\newblock
\showISBNx{978-1-66540-032-9}
\urldef\tempurl%
\url{https://doi.org/10.1109/IISA52424.2021.9555522}
\showDOI{\tempurl}


\bibitem[Marrel et~al\mbox{.}(2008)]%
        {marrel2008efficient}
\bibfield{author}{\bibinfo{person}{Amandine Marrel}, \bibinfo{person}{Bertrand
  Iooss}, \bibinfo{person}{Fran{\c{c}}ois Van~Dorpe}, {and}
  \bibinfo{person}{Elena Volkova}.} \bibinfo{year}{2008}\natexlab{}.
\newblock \showarticletitle{An efficient methodology for modeling complex
  computer codes with Gaussian processes}.
\newblock \bibinfo{journal}{\emph{Computational Statistics \& Data Analysis}}
  \bibinfo{volume}{52}, \bibinfo{number}{10} (\bibinfo{year}{2008}),
  \bibinfo{pages}{4731--4744}.
\newblock


\bibitem[{Martinez-Cantin} et~al\mbox{.}(2009)]%
        {martinez-cantin_bayesian_2009}
\bibfield{author}{\bibinfo{person}{Ruben {Martinez-Cantin}},
  \bibinfo{person}{Nando {de Freitas}}, \bibinfo{person}{Eric Brochu},
  \bibinfo{person}{Jos{\'e} Castellanos}, {and} \bibinfo{person}{Arnaud
  Doucet}.} \bibinfo{year}{2009}\natexlab{}.
\newblock \showarticletitle{A {{Bayesian}} Exploration-Exploitation Approach
  for Optimal Online Sensing and Planning with a Visually Guided Mobile Robot}.
\newblock \bibinfo{journal}{\emph{Autonomous Robots}} \bibinfo{volume}{27},
  \bibinfo{number}{2} (\bibinfo{date}{Aug.} \bibinfo{year}{2009}),
  \bibinfo{pages}{93--103}.
\newblock
\showISSN{1573-7527}
\urldef\tempurl%
\url{https://doi.org/10.1007/s10514-009-9130-2}
\showDOI{\tempurl}


\bibitem[Mockus(2012)]%
        {mockus2012bayesian}
\bibfield{author}{\bibinfo{person}{Jonas Mockus}.}
  \bibinfo{year}{2012}\natexlab{}.
\newblock \bibinfo{booktitle}{\emph{Bayesian approach to global optimization:
  theory and applications}}. Vol.~\bibinfo{volume}{37}.
\newblock \bibinfo{publisher}{Springer Science \& Business Media}.
\newblock


\bibitem[Nguyen(2019)]%
        {nguyen_bayesian_2019}
\bibfield{author}{\bibinfo{person}{Vu Nguyen}.}
  \bibinfo{year}{2019}\natexlab{}.
\newblock \showarticletitle{Bayesian {{Optimization}} for {{Accelerating
  Hyper-Parameter Tuning}}}. In \bibinfo{booktitle}{\emph{2019 {{IEEE Second
  International Conference}} on {{Artificial Intelligence}} and {{Knowledge
  Engineering}} ({{AIKE}})}}. \bibinfo{pages}{302--305}.
\newblock
\urldef\tempurl%
\url{https://doi.org/10.1109/AIKE.2019.00060}
\showDOI{\tempurl}


\bibitem[Nocedal and Wright(1999)]%
        {nocedal1999numerical}
\bibfield{author}{\bibinfo{person}{Jorge Nocedal} {and}
  \bibinfo{person}{Stephen~J Wright}.} \bibinfo{year}{1999}\natexlab{}.
\newblock \bibinfo{booktitle}{\emph{Numerical optimization}}.
\newblock \bibinfo{publisher}{Springer}.
\newblock


\bibitem[Papenmeier et~al\mbox{.}(2022)]%
        {papenmeier2022increasing}
\bibfield{author}{\bibinfo{person}{Leonard Papenmeier}, \bibinfo{person}{Luigi
  Nardi}, {and} \bibinfo{person}{Matthias Poloczek}.}
  \bibinfo{year}{2022}\natexlab{}.
\newblock \showarticletitle{Increasing the scope as you learn: Adaptive
  Bayesian optimization in nested subspaces}.
\newblock \bibinfo{journal}{\emph{Advances in Neural Information Processing
  Systems}}  \bibinfo{volume}{35} (\bibinfo{year}{2022}),
  \bibinfo{pages}{11586--11601}.
\newblock


\bibitem[Raponi et~al\mbox{.}(2019)]%
        {raponi_kriging-assisted_2019}
\bibfield{author}{\bibinfo{person}{Elena Raponi}, \bibinfo{person}{Mariusz
  Bujny}, \bibinfo{person}{Markus Olhofer}, \bibinfo{person}{Nikola Aulig},
  \bibinfo{person}{Simonetta Boria}, {and} \bibinfo{person}{Fabian Duddeck}.}
  \bibinfo{year}{2019}\natexlab{}.
\newblock \showarticletitle{Kriging-Assisted Topology Optimization of Crash
  Structures}.
\newblock \bibinfo{journal}{\emph{Computer Methods in Applied Mechanics and
  Engineering}}  \bibinfo{volume}{348} (\bibinfo{date}{May}
  \bibinfo{year}{2019}), \bibinfo{pages}{730--752}.
\newblock
\showISSN{0045-7825}
\urldef\tempurl%
\url{https://doi.org/10.1016/j.cma.2019.02.002}
\showDOI{\tempurl}


\bibitem[Raponi et~al\mbox{.}(2021)]%
        {raponi_methodology_2021}
\bibfield{author}{\bibinfo{person}{Elena Raponi}, \bibinfo{person}{Dario
  Fiumarella}, \bibinfo{person}{Simonetta Boria}, \bibinfo{person}{Alessandro
  Scattina}, {and} \bibinfo{person}{Giovanni Belingardi}.}
  \bibinfo{year}{2021}\natexlab{}.
\newblock \showarticletitle{Methodology for Parameter Identification on a
  Thermoplastic Composite Crash Absorber by the {{Sequential Response Surface
  Method}} and {{Efficient Global Optimization}}}.
\newblock \bibinfo{journal}{\emph{Composite Structures}} (\bibinfo{date}{Sept.}
  \bibinfo{year}{2021}), \bibinfo{pages}{114646}.
\newblock
\showISSN{0263-8223}
\urldef\tempurl%
\url{https://doi.org/10.1016/j.compstruct.2021.114646}
\showDOI{\tempurl}


\bibitem[Raponi et~al\mbox{.}(2020)]%
        {raponiHighDimensionalBayesian2020a}
\bibfield{author}{\bibinfo{person}{Elena Raponi}, \bibinfo{person}{Hao Wang},
  \bibinfo{person}{Mariusz Bujny}, \bibinfo{person}{Simonetta Boria}, {and}
  \bibinfo{person}{Carola Doerr}.} \bibinfo{year}{2020}\natexlab{}.
\newblock \showarticletitle{High Dimensional Bayesian Optimization Assisted by
  Principal Component Analysis}. In \bibinfo{booktitle}{\emph{Proc. of Parallel
  Problem Solving from Nature (PPSN)}} \emph{(\bibinfo{series}{LNCS},
  Vol.~\bibinfo{volume}{12269})}. \bibinfo{publisher}{Springer},
  \bibinfo{pages}{169--183}.
\newblock
\urldef\tempurl%
\url{https://doi.org/10.1007/978-3-030-58112-1\_12}
\showDOI{\tempurl}


\bibitem[Regis(2016)]%
        {regis2016trust}
\bibfield{author}{\bibinfo{person}{Rommel~G Regis}.}
  \bibinfo{year}{2016}\natexlab{}.
\newblock \showarticletitle{Trust regions in Kriging-based optimization with
  expected improvement}.
\newblock \bibinfo{journal}{\emph{Engineering optimization}}
  \bibinfo{volume}{48}, \bibinfo{number}{6} (\bibinfo{year}{2016}),
  \bibinfo{pages}{1037--1059}.
\newblock


\bibitem[Salem et~al\mbox{.}(2019)]%
        {salem2019sequential}
\bibfield{author}{\bibinfo{person}{Malek~Ben Salem},
  \bibinfo{person}{Fran{\c{c}}ois Bachoc}, \bibinfo{person}{Olivier Roustant},
  \bibinfo{person}{Fabrice Gamboa}, {and} \bibinfo{person}{Lionel Tomaso}.}
  \bibinfo{year}{2019}\natexlab{}.
\newblock \showarticletitle{Sequential dimension reduction for learning
  features of expensive black-box functions}.
\newblock  (\bibinfo{year}{2019}).
\newblock


\bibitem[Simionescu et~al\mbox{.}(2006)]%
        {simionescu2006two}
\bibfield{author}{\bibinfo{person}{Petru-Aurelian Simionescu},
  \bibinfo{person}{Gerry~Vernon Dozier}, {and} \bibinfo{person}{Roger~L
  Wainwright}.} \bibinfo{year}{2006}\natexlab{}.
\newblock \showarticletitle{A two-population evolutionary algorithm for
  constrained optimization problems}. In \bibinfo{booktitle}{\emph{2006 IEEE
  International Conference on Evolutionary Computation}}. IEEE,
  \bibinfo{pages}{1647--1653}.
\newblock


\bibitem[Sobester et~al\mbox{.}(2008)]%
        {sobesterEngineeringDesignSurrogate2008a}
\bibfield{author}{\bibinfo{person}{Andr{\'a}s Sobester},
  \bibinfo{person}{Alexander Forrester}, {and} \bibinfo{person}{Andy Keane}.}
  \bibinfo{year}{2008}\natexlab{}.
\newblock \bibinfo{booktitle}{\emph{Engineering {{Design}} via {{Surrogate
  Modelling}}: {{A Practical Guide}}}}.
\newblock \bibinfo{publisher}{{John Wiley \& Sons}}.
\newblock
\showISBNx{978-0-470-77079-5}


\bibitem[Turner et~al\mbox{.}(2021)]%
        {turner_bayesian_2021-2}
\bibfield{author}{\bibinfo{person}{Ryan Turner}, \bibinfo{person}{David
  Eriksson}, \bibinfo{person}{Michael McCourt}, \bibinfo{person}{Juha Kiili},
  \bibinfo{person}{Eero Laaksonen}, \bibinfo{person}{Zhen Xu}, {and}
  \bibinfo{person}{Isabelle Guyon}.} \bibinfo{year}{2021}\natexlab{}.
\newblock \bibinfo{title}{Bayesian {{Optimization}} Is {{Superior}} to {{Random
  Search}} for {{Machine Learning Hyperparameter Tuning}}: {{Analysis}} of the
  {{Black-Box Optimization Challenge}} 2020}.
\newblock
\newblock
\urldef\tempurl%
\url{https://doi.org/10.48550/arXiv.2104.10201}
\showDOI{\tempurl}
\showeprint[arxiv]{2104.10201}~[cs, stat]


\bibitem[Vikhar(2016)]%
        {vikhar2016evolutionary}
\bibfield{author}{\bibinfo{person}{Pradnya~A Vikhar}.}
  \bibinfo{year}{2016}\natexlab{}.
\newblock \showarticletitle{Evolutionary algorithms: A critical review and its
  future prospects}. In \bibinfo{booktitle}{\emph{2016 International conference
  on global trends in signal processing, information computing and
  communication (ICGTSPICC)}}. IEEE, \bibinfo{pages}{261--265}.
\newblock


\bibitem[Wang et~al\mbox{.}(2022)]%
        {IOHprofiler}
\bibfield{author}{\bibinfo{person}{Hao Wang}, \bibinfo{person}{Diederick
  Vermetten}, \bibinfo{person}{Furong Ye}, \bibinfo{person}{Carola Doerr},
  {and} \bibinfo{person}{Thomas B{\"a}ck}.} \bibinfo{year}{2022}\natexlab{}.
\newblock \showarticletitle{{{IOHanalyzer}}: {{Detailed Performance Analyses}}
  for {{Iterative Optimization Heuristics}}}.
\newblock \bibinfo{journal}{\emph{{ACM} Trans. Evol. Learn. Optim.}}
  \bibinfo{volume}{2}, \bibinfo{number}{1} (\bibinfo{year}{2022}),
  \bibinfo{pages}{3:1--3:29}.
\newblock
\showISSN{2688-299X}
\urldef\tempurl%
\url{https://doi.org/10.1145/3510426}
\showDOI{\tempurl}


\bibitem[Wang et~al\mbox{.}(2018)]%
        {wang2018batched}
\bibfield{author}{\bibinfo{person}{Zi Wang}, \bibinfo{person}{Clement Gehring},
  \bibinfo{person}{Pushmeet Kohli}, {and} \bibinfo{person}{Stefanie Jegelka}.}
  \bibinfo{year}{2018}\natexlab{}.
\newblock \showarticletitle{Batched large-scale Bayesian optimization in
  high-dimensional spaces}. In \bibinfo{booktitle}{\emph{International
  Conference on Artificial Intelligence and Statistics}}. PMLR,
  \bibinfo{pages}{745--754}.
\newblock


\bibitem[Wang et~al\mbox{.}(2016)]%
        {wang_bayesian_2016}
\bibfield{author}{\bibinfo{person}{Ziyu Wang}, \bibinfo{person}{Frank Hutter},
  \bibinfo{person}{Masrour Zoghi}, \bibinfo{person}{David Matheson}, {and}
  \bibinfo{person}{Nando {de Freitas}}.} \bibinfo{year}{2016}\natexlab{}.
\newblock \showarticletitle{Bayesian {{Optimization}} in a {{Billion
  Dimensions}} via {{Random Embeddings}}}.
\newblock \bibinfo{journal}{\emph{arXiv:1301.1942}} (\bibinfo{year}{2016}).
\newblock


\bibitem[Wycoff et~al\mbox{.}(2019)]%
        {wycoff2019sequential}
\bibfield{author}{\bibinfo{person}{Nathan Wycoff}, \bibinfo{person}{Mickael
  Binois}, {and} \bibinfo{person}{Stefan~M Wild}.}
  \bibinfo{year}{2019}\natexlab{}.
\newblock \showarticletitle{Sequential learning of active subspaces}.
\newblock \bibinfo{journal}{\emph{arXiv preprint arXiv:1907.11572}}
  (\bibinfo{year}{2019}).
\newblock


\bibitem[Zhigljavsky and {\v{Z}}ilinskas(2021)]%
        {zhigljavsky2021bayesian}
\bibfield{author}{\bibinfo{person}{Anatoly Zhigljavsky} {and}
  \bibinfo{person}{Antanas {\v{Z}}ilinskas}.} \bibinfo{year}{2021}\natexlab{}.
\newblock \bibinfo{booktitle}{\emph{Bayesian and high-dimensional global
  optimization}}.
\newblock \bibinfo{publisher}{Springer}.
\newblock


\bibitem[Zhigljavsky(2012)]%
        {zhigljavsky2012theory}
\bibfield{author}{\bibinfo{person}{Anatoly~A Zhigljavsky}.}
  \bibinfo{year}{2012}\natexlab{}.
\newblock \bibinfo{booktitle}{\emph{Theory of global random search}}.
  Vol.~\bibinfo{volume}{65}.
\newblock \bibinfo{publisher}{Springer Science \& Business Media}.
\newblock


\bibitem[Ziomek and Bou~Ammar(2023)]%
        {pmlr-v202-ziomek23a}
\bibfield{author}{\bibinfo{person}{Juliusz~Krzysztof Ziomek} {and}
  \bibinfo{person}{Haitham Bou~Ammar}.} \bibinfo{year}{2023}\natexlab{}.
\newblock \showarticletitle{Are Random Decompositions all we need in High
  Dimensional {B}ayesian Optimisation?}. In
  \bibinfo{booktitle}{\emph{Proceedings of the 40th International Conference on
  Machine Learning}} \emph{(\bibinfo{series}{Proceedings of Machine Learning
  Research}, Vol.~\bibinfo{volume}{202})},
  \bibfield{editor}{\bibinfo{person}{Andreas Krause}, \bibinfo{person}{Emma
  Brunskill}, \bibinfo{person}{Kyunghyun Cho}, \bibinfo{person}{Barbara
  Engelhardt}, \bibinfo{person}{Sivan Sabato}, {and} \bibinfo{person}{Jonathan
  Scarlett}} (Eds.). \bibinfo{publisher}{PMLR}, \bibinfo{pages}{43347--43368}.
\newblock
\urldef\tempurl%
\url{https://proceedings.mlr.press/v202/ziomek23a.html}
\showURL{%
\tempurl}


\end{thebibliography}

%%%%%%%%%%%%%%%%%%%%
%%%%%%%%%%%%%%%%%%%%
%%%%%%%%%%%%%%%%%%%%
%%%%%%%%%%%%%%%%%%%%

\newpage
\onecolumn
\appendix

\section*{Appendix}
 \vspace{.5cm}
\section{Algorithms}
\label{sec:algo}
Here we provide the reader with a more detailed description of the algorithms that were compared in our study and the chosen hyperparameter settings.
\subsection{Covariance Matrix Adaptation Evolution Strategy (CMA-ES)} \label{sec:Covariance Matrix Adaptation Evolution Strategy}
\textbf{Covariance Matrix Adaptation Evolution Strategy (CMA-ES)})~\cite{hansen2006cma, hansen2004evaluating, hansen2003reducing} belongs to the family of evolutionary algorithms~\cite{vikhar2016evolutionary, simionescu2006two} and is considered to be the state-of-the-art in this category, being the most commonly used for continuous optimization in many research laboratories and industrial environments. CMA-ES is mostly used to solve difficult nonlinear nonconvex black-box optimization problems, unconstrained or constrained, in continuous domains. It efficiently addresses search spaces of dimension between three and one hundred. CMA-ES is based on an idea similar to the Quasi-Newton method~\cite{nocedal1999numerical}, that is, it is a second-order estimator that estimates a positive definite matrix, the covariance matrix, in an iterative way. Unlike the Quasi-Newton method, CMA-ES does not use or approximate gradients and does not even assume their existence. For this reason, it can be used for non-smooth and even non-continuous problems, as well as for multimodal and/or noisy problems.
Furthermore, CMA-ES does not require expensive hyperparameter tuning, since the choice of hyperparameters is not left to the user (apart from population size). The only hyperparameters that the user needs to set for the application of CMA-ES are an initial parent design, an initial step-size, and a termination criterion.

In this study, the initial solution was taken as a random sample array in the design space and the initial step-size in each coordinate is equal to 1. We considered a default population size of $4 + 3\log D$, where $\log D$ is the natural logarithm of the dimension of the design domain.

\subsection{Sparse Axis-Aligned Subspaces BO (SAASBO)} \label{sec:Sparse Axis-Aligned Subspaces BO}

One of the most difficult problems with high-dimensional BO is defining an appropriate class of surrogate models. On the one hand, a class of models that is too flexible may lead to overfitting.
On the other hand, a class that is too rigid would not be able to capture the important properties of the objective function landscape. 

\textbf{Sparse Axis Aligned Subspace Bayesian Optimization (SAASBO)}~\cite{eriksson2021high} introduces a new surrogate model for high-dimensional BO based on the assumption that the coordinates of $ x \text{ in } S$ have a relevance hierarchy.
According to~\cite{eriksson2021high} this approach has several key advantages. First, it preserves the structure of the input space and therefore it can exploit it. Second, it is adaptive and shows low sensitivity to its hyperparameters. Third, it can naturally accommodate both input and output constraints, unlike methods based on random projections for which input constraints are particularly challenging.
In SAASBO, the usual Gaussian process is used, but with the help of some new components.

The innovations are:
\begin{itemize}
    \item sparsity-inducing SAAS function prior;
    \item combination of the surrogate model with the No-Turn-U-Sampler (NUTS), an adaptive form of Hamiltonian Monte Carlo (HMC) sampling that the algorithm must perform to do inference on that model. It allows the surrogate model to quickly identify the most important low-dimensional subspace, resulting in a sample-efficient BO.
\end{itemize}

 In our experiments, we set the following values for the main hyperparameters:
\begin{itemize}
    \item $\alpha = 0.1$, a positive float that controls the level of shrinkage/sparsity of the GP model. Smaller alpha for more sparsity, and so most dimensions “turned off”;
    \item num\_warmup = 256, the number of warmup samples to use in the NUTS. During the warm-up, the NUTS algorithm adjusts the HMC algorithm parameters metric and step-size in order to sample efficiently. After the warm-up, the fixed metric and step-size are used to produce a set of draws;
    \item num\_samples = 256, the number of post-warmup samples to use in HMC inference;
    \item thinning = 32, a positive integer that controls the fraction of posterior hyperparameter samples that are used to compute the expected improvement;
    \item kernel = rbf. By default, SAASBO uses radial basis function kernels in the GP model definition.
\end{itemize}

\subsection{Random Decomposition Upper-Confidence Bound (RDUCB)}
\label{sec:RDUCB}
\textbf{Random Decomposition Upper-Confidence Bound (RDUCB)}~\cite{pmlr-v202-ziomek23a} introduces a decomposition method that assumes additively structured black boxes based on data-independent decomposition rules. The reason behind this is that learning the decomposition of the function based on the data can be misleading if the dataset is not fixed, but rather data is dynamically added during the iterations. Indeed, data-driven approaches can easily fail to apply globally to the entire search space and lead to erroneous local decompositions.

RDUCB is an easy-to-implement approach that leads to empirical gains while supporting rigorous theoretical results. RDUCB uses an additive approach with a tree structure. This method efficiently achieves accurate approximations by randomly sampling trees, ensuring each edge has an equal chance of being selected. The acquisition function to be maximized is an additive version of the upper-confidence bound function. They achieve this through the utilization of message-passing optimizers, which can more effectively leverage the knowledge of the dependency graph. This approach results in improved empirical performance by transmitting information from children to parent nodes of the tree.

In our experiments we set the following values for the main hyperparameters:
\begin{itemize}
    \item eps = $-1$, minimum distance between two consecutive x-values to keep running the model;
    \item exploration\_weight = 'lambda t: 0.5 np.log(2t)', a configurable hyperparameter of the acquisition function that controls the exploration behavior, where t denotes the timestep;
    \item graphSamplingNumIter = 100, number of iterations to build the best graph;
    \item learnDependencyStructureRate = 1, hyperparameter that controls when to learn a new GP model;
    \item lengthscaleNumIter = 2, number of samples taken for the parameters during the optimization of the graph;
    \item max\_eval = -4, hyperparameter of the optimizer of the acquisition function that decides how many times to divide the search space into subdomains to find a better solution;
    \item noise\_var = 0.1, observation error of the GP model;
    \item size\_of\_random\_graph = 0.2, to decide the dimension of the random graph;
    \item grid\_size = 150, which controls the density of the discretization, given that the domain is a discretized representation of the search space.
\end{itemize}

\subsection{Linear PCA-assisted Bayesian
Optimization (PCA-BO)} \label{sec:Linear PCA-assisted Bayesian Optimization}

Another method for scaling BO for high-dimensional data is based on the use of Principal Component Analysis (PCA) to generate a new BO algorithm called \textbf{PCA-assisted Bayesian optimization (PCA-BO)}~\cite{raponiHighDimensionalBayesian2020a}. Thanks to PCA, the algorithm learns a linear transformation based on the points evaluated so far and selects the dimensions in the transformed space considering the variability of these points. Both the fitting of the GPR model and the optimization of the acquisition function are carried out in the space with reduced dimensionality. The primary benefits are the reduction of CPU time for high-dimensional problems and the ability to maintain a good convergence rate for problems with an adequate global structure.

PCA-BO starts with a DoE and adaptively learns a linear map, updated at each iteration, to reduce the dimensionality through a weighted PCA procedure. Weights are used to account for objective values.

The main steps of the PCA-BO algorithm can be summarized as follows:
\begin{enumerate}
    \item Generate an initial DoE in the original space composed of a set of evenly distributed points;
    \item Design a weighted scheme to take into account the information from the objective function into the DoE points: smaller weights are assigned to the points with worse function values.
    \item Apply the PCA technique to create a linear map from the original search space to a lower-dimensional space, using the weighted DoE points;
    \item Train a GPR model and maximize an infill criterion to find the new infill point, both in the lower-dimensional space;
    \item Map back the infill point to the original space and evaluate its objective function value;
    \item Append the new infill point and its objective value to the data set and then proceed to Step 2 for a new iteration.
\end{enumerate}

 We set the following values for the main hyperparameters:
\begin{itemize}
    \item n\_point = 1, the number of infill points selected during the optimization of the acquisition function;
    \item n\_components = 0.90, the amount of variance that needs to be explained by the principal components kept after dimensionality reduction by PCA;
    \item acquisition\_optimization = BFGS, optimizer used to maximize the acquisition function. 
\end{itemize}

\subsection{Kernel PCA-assisted BO (KPCA-BO)} \label{sec:Kernel PCA-assisted BO}
The \textbf{kernel PCA-assisted algorithm BO (KPCA-BO)}~\cite{kPCABO} is an extended version of PCA-BO, which uses kernel methods to first map points to a reproducing kernel Hilbert space (RKHS)~\cite{berlinet2011reproducing} using an implicit nonlinear feature mapping. Then, PCA is used in the RKHS to learn a linear transformation from all evaluated points during the run and select dimensions in the transformed space according to the variability of the evaluated points. The main advantage of KPCA-BO over PCA-BO is the nonlinearity of the submanifold in the search space, which makes it more likely that multiple basins of attraction will be discovered simultaneously.
Besides learning a forward map from the original space to a lower-dimensional submanifold, it constructs a backward map that converts an infill point determined in the reduced space to the original space, so that it can be evaluated. In this way, the two most extensive procedures of BO, training the GP model and optimizing the acquisition function, are performed in a low-dimensional space, reducing CPU time.

We set the following values for the main hyperparameters:
\begin{itemize}
    \item n\_point = 1, the number of infill points selected during the optimization of the acquisition function;
    \item max\_information\_loss=0.1, the decimal value of the maximum information of variability data points that can be lost during the PCA procedure in the Hilbert space;
    \item kernel\_fit\_strategy = KernelFitStrategy.AUTO. The AUTO setting uses an RBF kernel.
\end{itemize}

\subsection{Trust Region BO (TuRBO)} \label{TURBO}
Two key issues often lead to poor performance of classical BO in high-dimensional settings: the homogeneity of global probabilistic models and overemphasized exploration due to global coverage. To overcome these problems, the global perspective can be discarded, and a local approach can be used.\textbf{Trust regions BO (TuRBO)}~\cite{eriksson2019scalable} is a technique for global optimization that uses a collection of simultaneous local optimization runs with independent probabilistic models. To combine all the local parts and return to a global vision, an implicit multi-armed bandit strategy is used at each iteration to distribute the samples across different local domains and hence decide which local optimization runs are preferred.
In each global iteration, which means considering all the trust regions, the algorithm selects a batch of $q$ candidates drawn from the union of all the trust regions, and updates all the local models for which candidates were drawn. A Thompson sampling~\cite{hernandez2016distributed} is used to select infill points within a single trust region and in all trust regions simultaneously. 

The main advantages of this approach are that (1) each local surrogate model is robust to noisy observations and uncertainty estimates, (2) the local surrogates allow heterogeneous modeling of the objective function and do not suffer from over-exploitation, and (3) it provides local search trajectories to quickly discover excellent target values.

We set the hyperparameters as follows:
\begin{itemize}
    \item batch\_size = 5, number of infill points found in each iteration,
    \item max\_cholesky\_size = 2000, after how many iterations the algorithm switches from Cholesky to Lanczos to train the GP;
    \item n\_training\_steps = 50, number of steps of ADAM to learn the hyperparameters of the GP;
    \item n\_cand = min(100 * D, 5000), number of vectors of dimension D generated by a Sobolev sequence where to evaluate the batch\_size GP samples;
    \item failtol = ceil(max([4.0/ batch\_size, D/batch\_size])) for TuRBO1 and failtol = max(5, D) for TuRBOm, where ceil of the scalar $x$ is the smallest integer $i$ such that $i \leq x$ and it is the threshold for failures after which trust regions halves;
    \item succtol = 3, the threshold for successes after which trust regions doubles;
    \item length\_min = $0.5^7$, the minimum threshold for the base side length of the trust regions before restart;
    \item length\_max = 1.6, the maximum threshold for the base side length of the trust regions;
    \item length\_init = 0.8, value to initialize the base side length of the trust regions.
\end{itemize}

\subsection{Ensemble BO (EBO)} \label{sec: Ensemble Bayesian Optimization}

In our data on Zenodo and IOHanalyzer, we also included results for \textbf{Ensemble Bayesian Optimization (EBO)}~\cite{wang2018batched}. This algorithm belongs to the additive models category.
When dealing with high-dimensional problems, the inefficiency of BO occurs not only in the creation of the surrogate model but also in the optimization of the acquisition function, which is sometimes very expensive to evaluate. Moreover, reliable search and estimation for complex functions in very high-dimensional spaces may require a large number of observations. EBO attempts to answer precisely these three challenges:
 \begin{enumerate}
     \item large-scale observations,
     \item high-dimensional input spaces,
     \item selection of batches of query points that balance quality and diversity.
\end{enumerate}

To solve the three challenges, the authors propose to improve the GP models by using a hierarchical additive approach based on tile coding, also known as random binning or Mondrian Forest features. Then, thanks to a Gibbs sampling, the posterior distribution is learned over the kernel width and the additive structure to prevent overfitting. The third challenge is to improve the sampler which depends on the likelihood of the observations. This is accelerated by an efficient randomized block approximator of the Gram matrix based on a Mondrian process.
Thus, EBO relies on two main ideas implemented at different levels: 1) using efficient partition-based function approximators to simplify and speed up the model-building and the optimization procedure and 2) improving the expressive power of these approximators by using ensembles and a stochastic approach that relies on the Mondrian process. Moreover, they use an ensemble of Tile Gaussian Processes (TileGPs) for each part, a new GP model based on tile coding and additive structure. Their method can be defined as a stochastic method over a randomized and adaptive sample of partitions of the input data space. 

The EBO code is taken from the GitHub repository \verb|Ensemble-Bayesian-Optimization|,\footnote{~\url{https://github.com/zi-w/Ensemble-Bayesian-Optimization}} but we redefine the acquisition function as the EI, because the default implementation uses a global minimum value that is assumed to be known, while we assume that it works in a complete black-box scenario. In our experiments, two different implementations of the EBO algorithm are utilized: EBO and EBO\_B. They differ for the value of the hyperparameter $B$, which represents the number of query points selected at each iteration. We use $B=1$ and $B=10$, respectively. To use the same total budget, EBO\_B runs for budget/10 iterations.
 
 We set the following values for the main hyperparameters:
\begin{itemize}
    \item z = sample\_z(D). This parameter controls the additive decomposition of the input feature space. It is an array of dimension D and it can assume only discrete values. Here it is selected randomly through the method \verb|sample_z|;
    \item k = array([10] * D), the number of cuts in each dimension. It is an array of dimension D and it can assume only discrete values;
    \item $\alpha = 1$, hyperparameter of the Gibbs sampling subroutine;
    \item $\beta = \text{array}([5.,2.])$, hyperparameter of the Gibbs sampling subroutine;
 \item opt\_n = 1000 points randomly sampled to initiate the continuous optimization of the acquisition function;
 \item gibbs\_iter = 10 number of iterations for the Gibbs sampling subroutine;
 \item nlayers = 100, number of the layers of tiles;
 \item gp\_sigma = 0.1, noise standard deviation;
 \item n\_top = 10, how many points to look ahead when selecting the new infill point.
\end{itemize}

The advantage we can observe is a low CPU time for optimizing the acquisition function, but this advantage cannot outweigh the poor convergence performance. In fact, stagnation at low budget is a very common behavior of this algorithm in the problems we treat. We attribute this to the choice of its default hyperparameters, which may not have been designed for our function landscapes. For these reasons, EBO was not further considered in great detail in our work, and we decided to use RDUCB~\cite{pmlr-v202-ziomek23a} as representative for the additive model approaches.

\section{Extensive CPU time analysis}
\label{sec:moreCPUtime}

\begin{figure}[ht]
\centering
\subfloat[Bar plot showing the CPU time in seconds for the acquisition function optimization phase function by function. Time is averaged over all iterations of one run.] {
	\label{subfig:acqtime10}
	\includegraphics[width=\columnwidth]{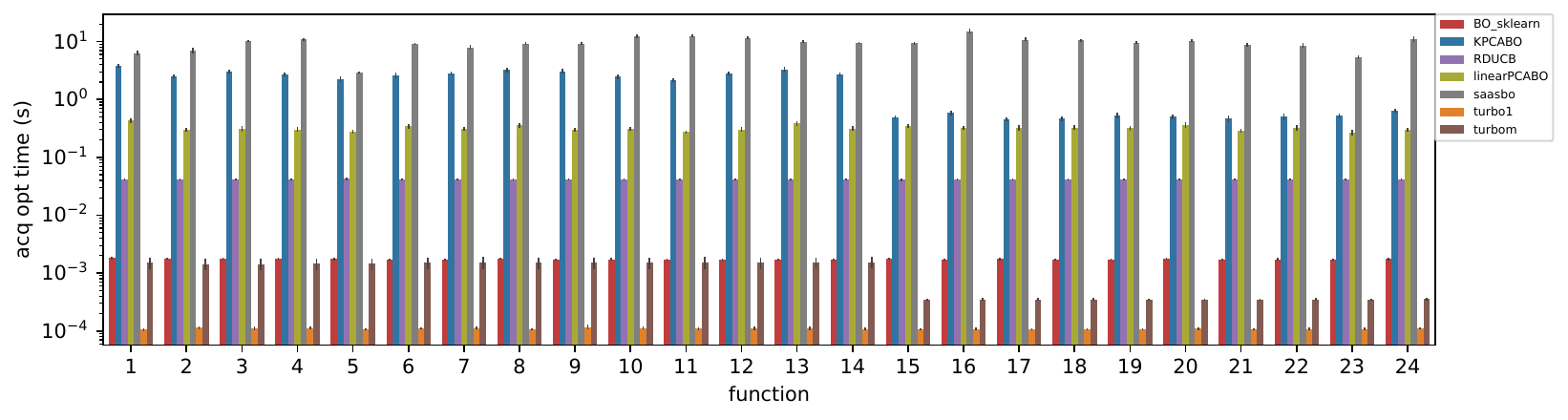} } 
	\hfill
\subfloat[Bar plot showing CPU time in seconds to fit the model, function by function. Time is averaged over all iterations of one run.]{
	\label{subfig:modeltime10}
	\includegraphics[width=\columnwidth]{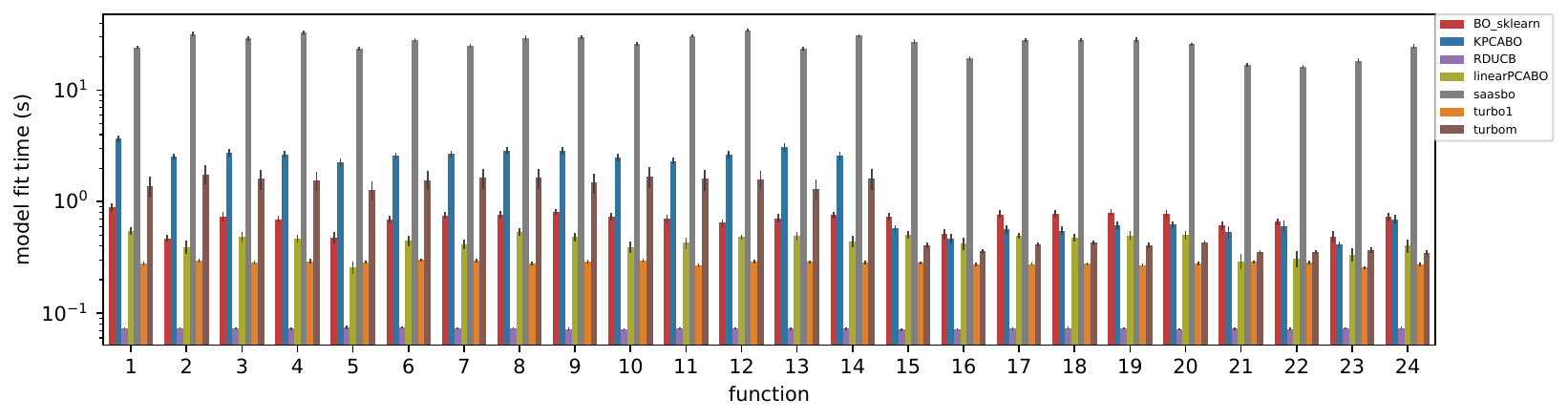} } 
		\hfill
\subfloat[Bar plot showing CPU time in seconds for a whole run, function by function.]{
	\label{subfig:cumtime10}
	\includegraphics[width=\columnwidth]{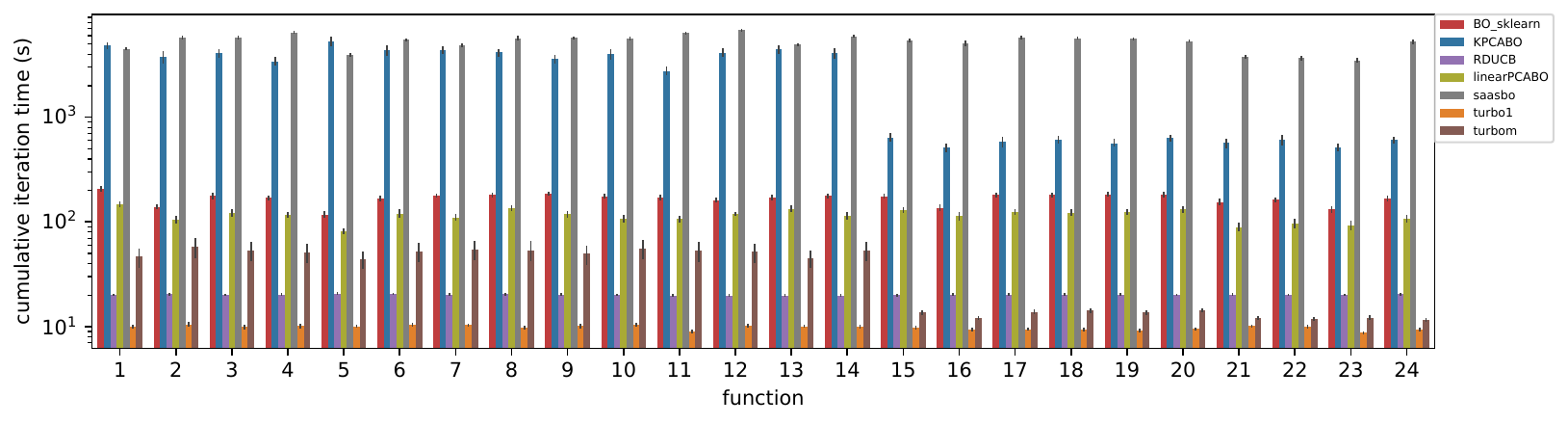} } 
\caption{CPU time bar plots for 10D, function by function.}
\label{10Dseparating}
\end{figure}

\begin{figure}[ht]
\centering
\subfloat[Bar plot showing the CPU time in seconds for the acquisition function optimization phase function by function. Time is averaged over all iterations of one run.] {
	\label{subfig:acqtime20}
	\includegraphics[width=\columnwidth]{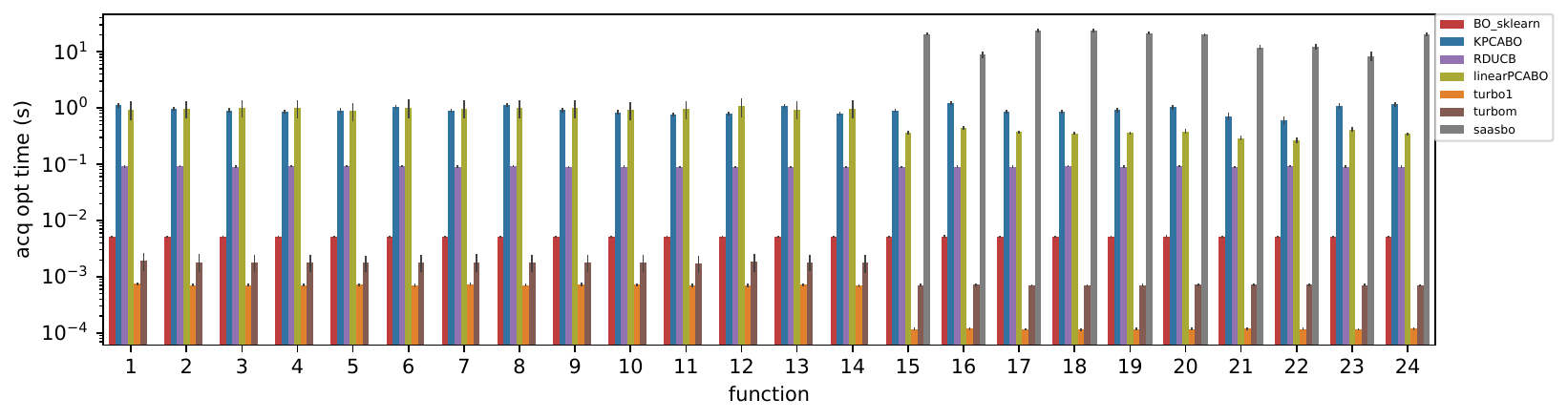} } 
	\hfill
\subfloat[Bar plot showing CPU time in seconds to fit model phase, function by function. Time is averaged over all iterations of one run.]{
	\label{subfig:modeltime20}
	\includegraphics[width=\columnwidth]{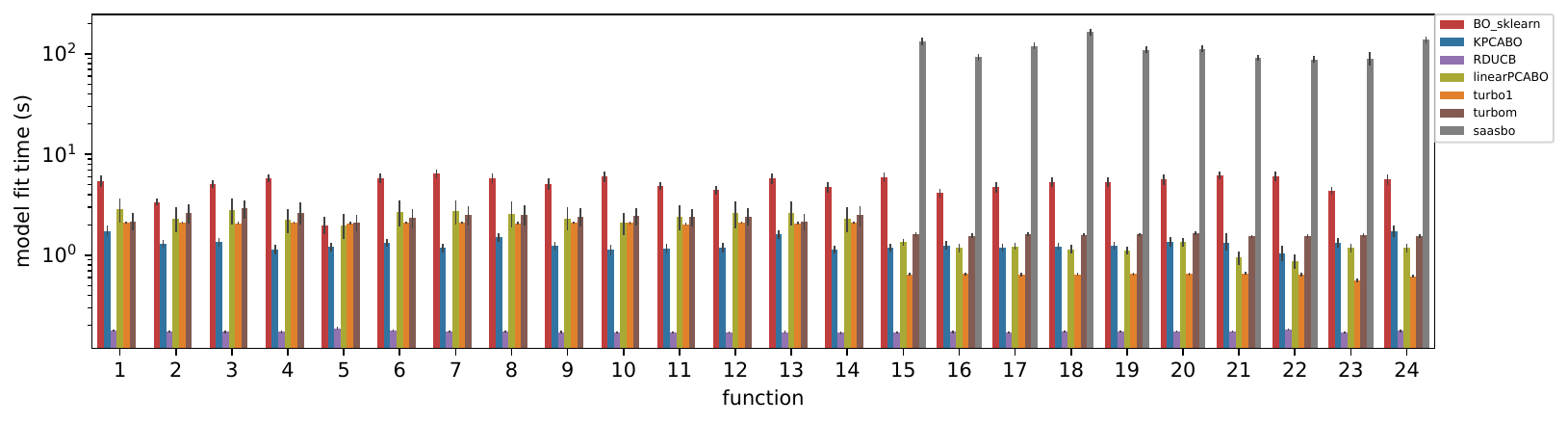} } 
		\hfill
\subfloat[Bar plot showing CPU time in seconds for a whole run, function by function.]{
	\label{subfig:cumtime20}
	\includegraphics[width=\columnwidth]{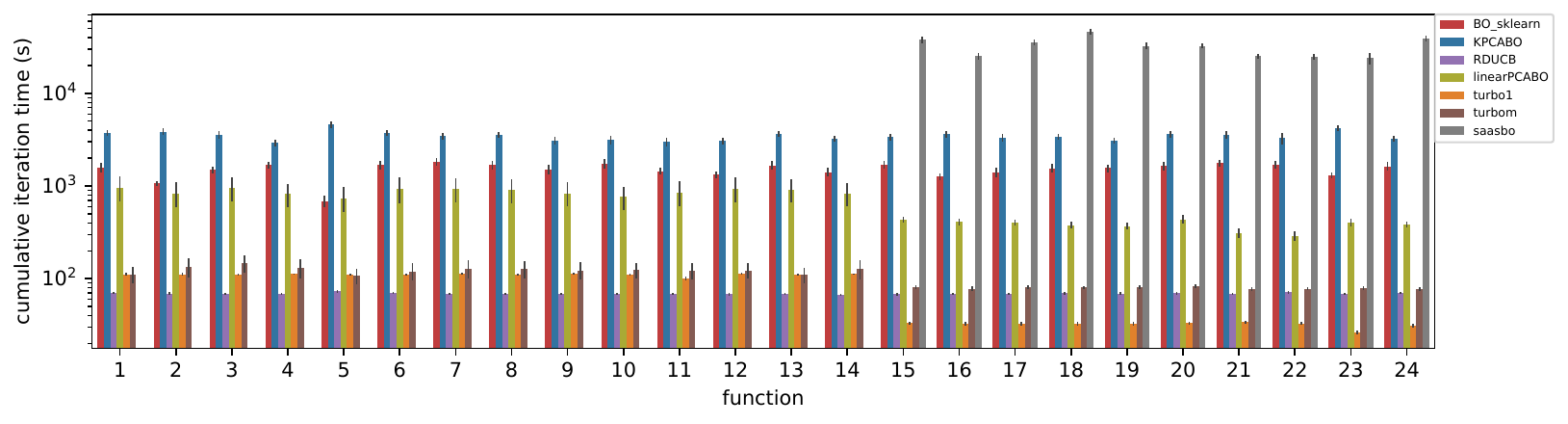} } 
\caption{CPU time bar plots for 20D, function by function.}
\label{20Dseparating}
\end{figure}

\begin{figure}[ht]
\centering
\subfloat[Bar plot showing the CPU time in seconds for the acquisition function optimization phase function by function. Time is averaged over all iterations of one run.]{
	\label{subfig:acqtime40}
	\includegraphics[width=\columnwidth]{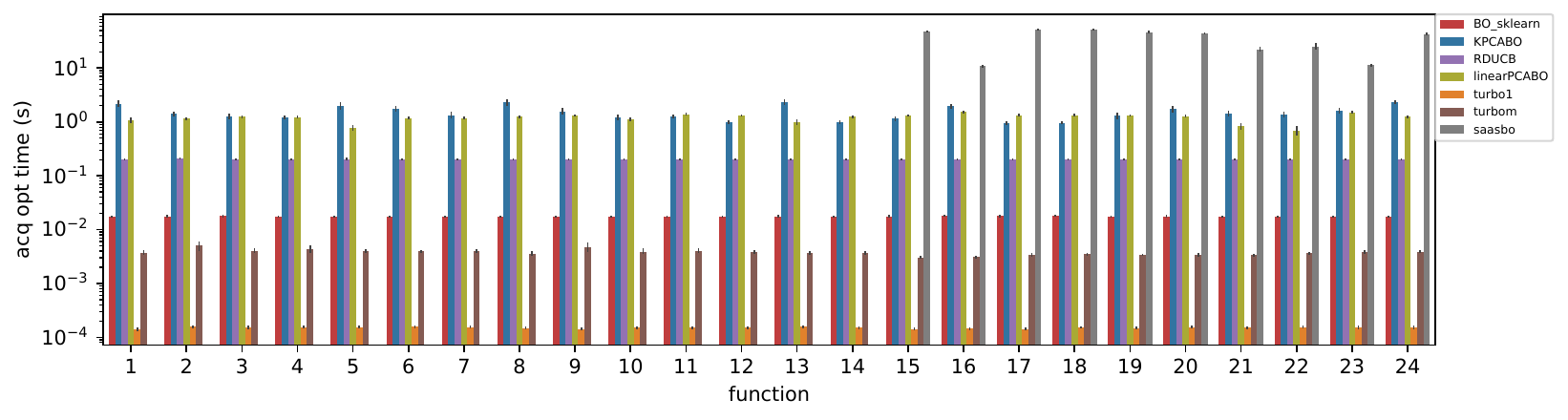} } 
	\hfill
\subfloat[Bar plot showing the CPU time in seconds for the fit of the model phase, function by function. Time is averaged over all iterations of one run.]{
	\label{subfig:modeltime40}
	\includegraphics[width=\columnwidth]{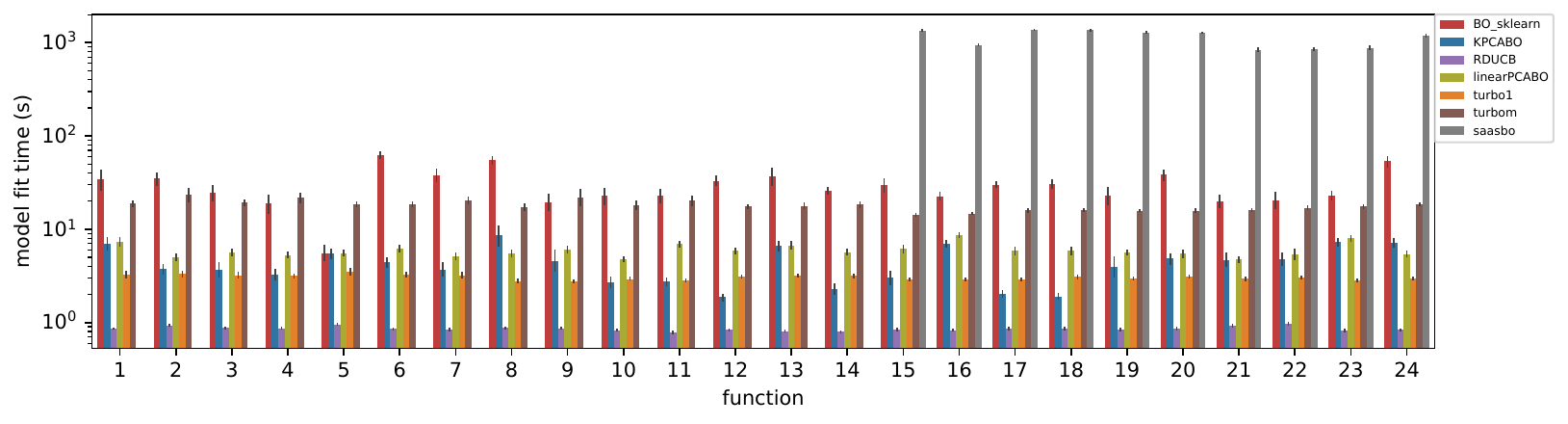} } 
		\hfill
\subfloat[Bar plot showing the CPU time in seconds for a whole run, function by function.]{
	\label{subfig:cumtime40}
	\includegraphics[width=\columnwidth]{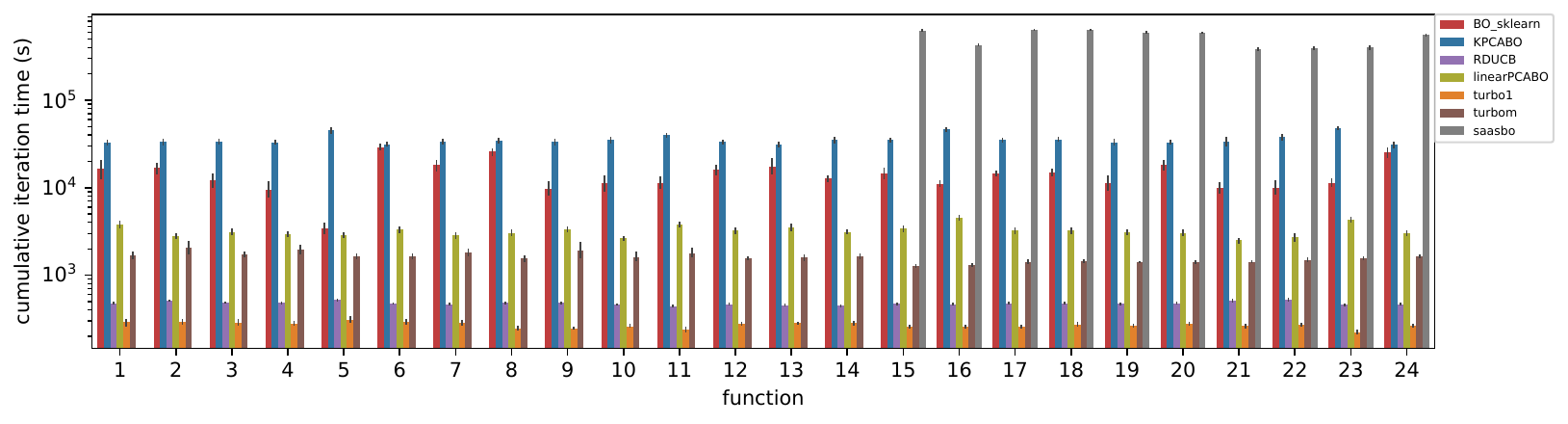} } 
\caption{CPU time bar plots for 40D, function by function.}
\label{40Dseparating}
\end{figure}

\begin{figure}[ht]
\centering
\subfloat[Bar plot showing the CPU time in seconds for the acquisition function optimization phase function by function. Time is averaged over all iterations of one run.]{
	\label{subfig:acqtime60}
	\includegraphics[width=\columnwidth]{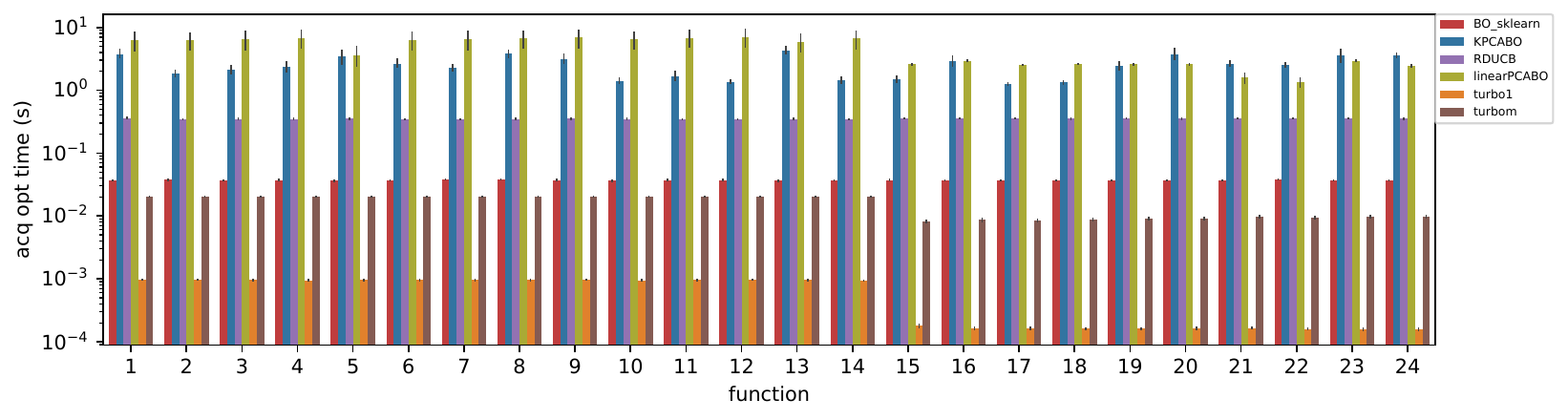} } 
	\hfill
\subfloat[Bar plot showing CPU time in seconds to fit model phase, function by function. Time is averaged over all iterations of one run.]{
	\label{subfig:modeltime60}
	\includegraphics[width=\columnwidth]{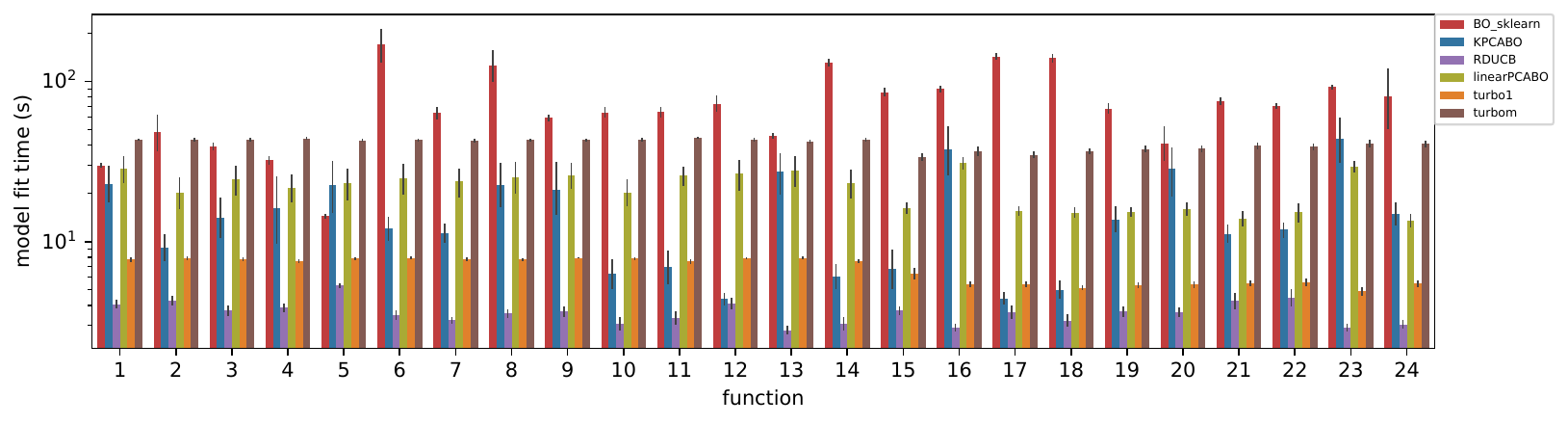} } 
		\hfill
\subfloat[Bar plot showing CPU time in seconds for the entire process, function by function.]{
	\label{subfig:cumtime60}
	\includegraphics[width=\columnwidth]{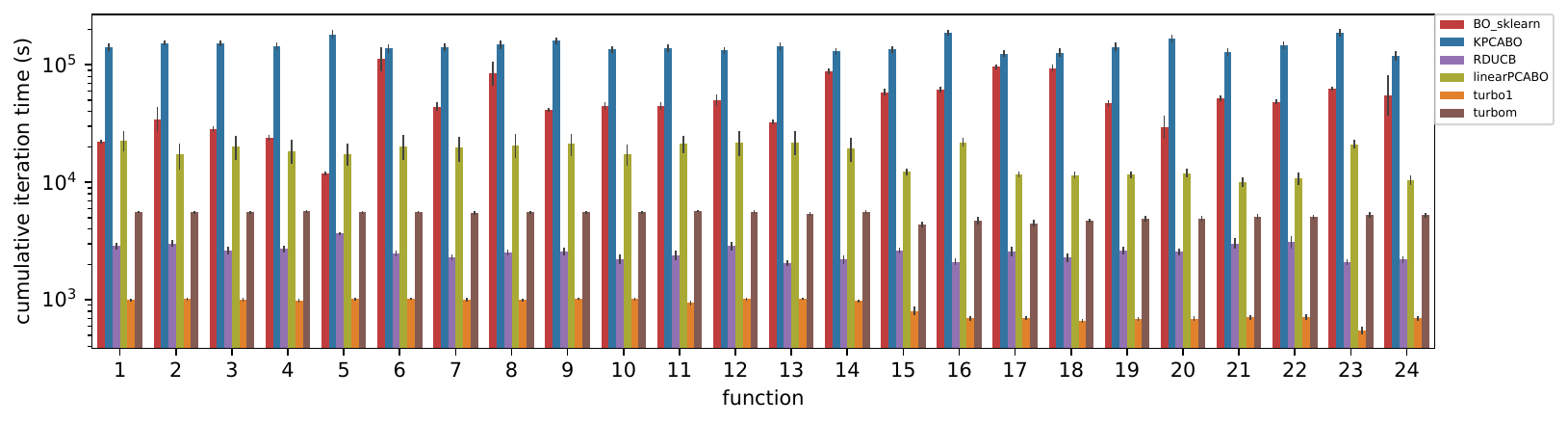} } 
\caption{CPU time bar plots for 60D, function by function.}
\label{60Dseparating}
\end{figure}

Figures~\ref{10Dseparating}, \ref{20Dseparating}, \ref{40Dseparating}, and~\ref{60Dseparating} present additional bar plots for the CPU time required to fit the model, optimize the acquisition function, and perform the complete run, for each method and on each BBOB function.

\clearpage
\section{Violin Plots}
\begin{figure}[ht] \center
    \includegraphics[width=\columnwidth]{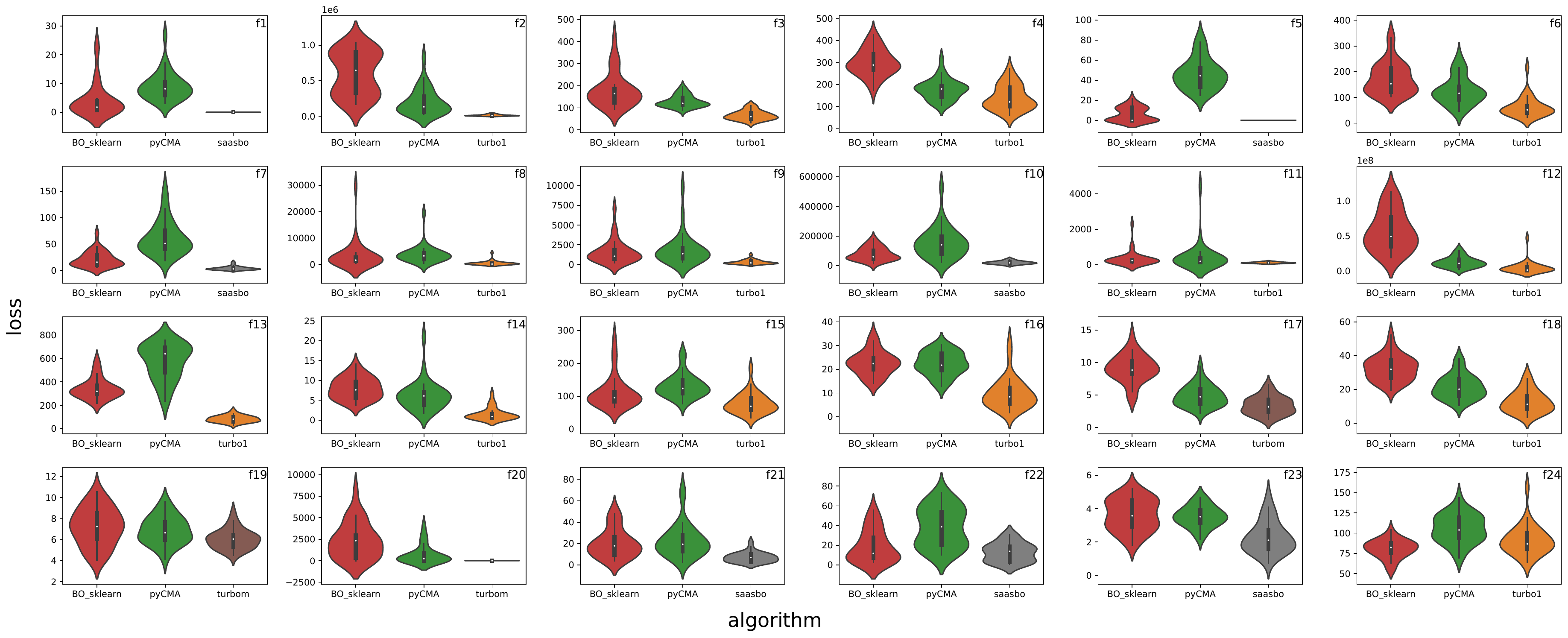}
      \caption{The violin plots show, for dimension 10 and budget 150 (final budget), the distribution of the best-so-far function values for all functions, obtained by vanilla BO, CMA-ES, and the best among the HDBO algorithms. The plots include a marker for the median of the data and a box indicating the interquartile range.}
    \label{boxplotD10}
\end{figure}

\begin{figure}[ht] \center
    \includegraphics[width=\columnwidth]{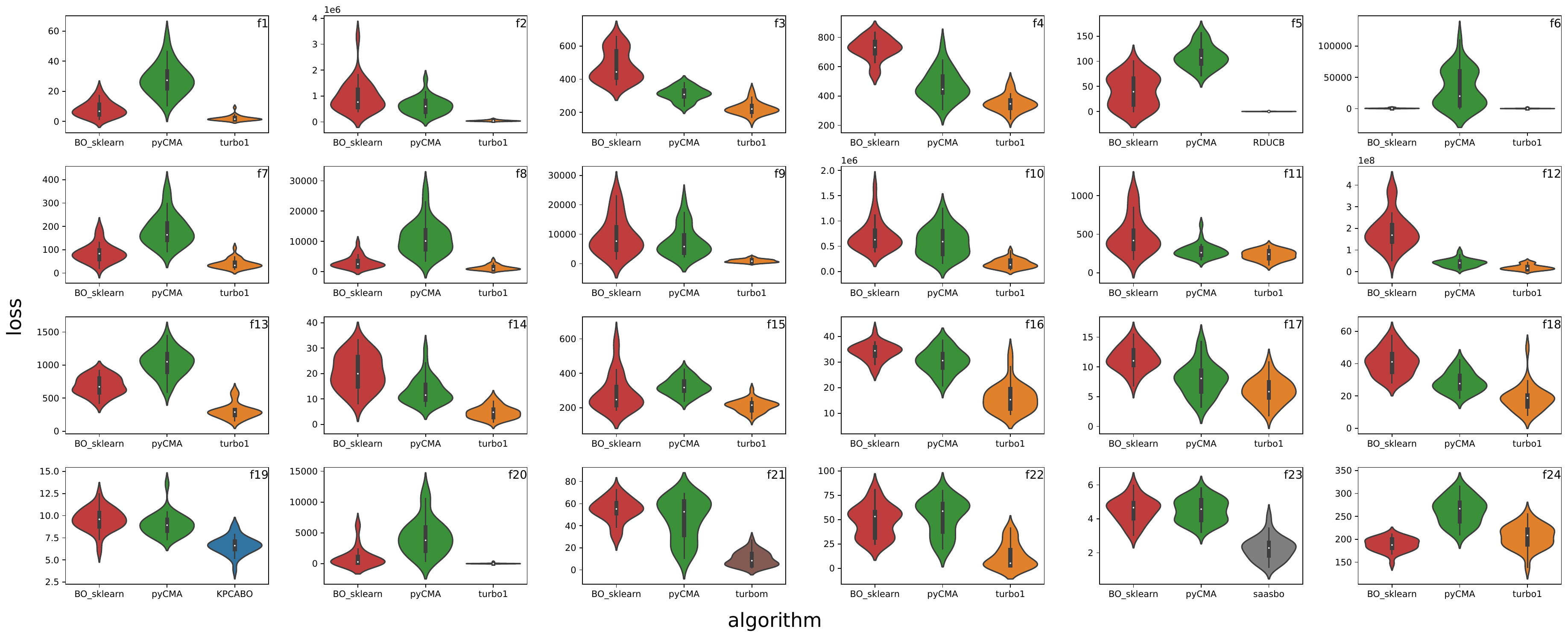}
      \caption{The violin plots show, for dimension 20 and budget 250 (final budget), the distribution of the best-so-far function values for all functions, obtained by vanilla BO, CMA-ES, and the best among the HDBO algorithms. The plots include a marker for the median of the data and a box indicating the interquartile range.}
    \label{boxplotD20}
\end{figure}

\begin{figure}[ht] \center
    \includegraphics[width=\columnwidth]{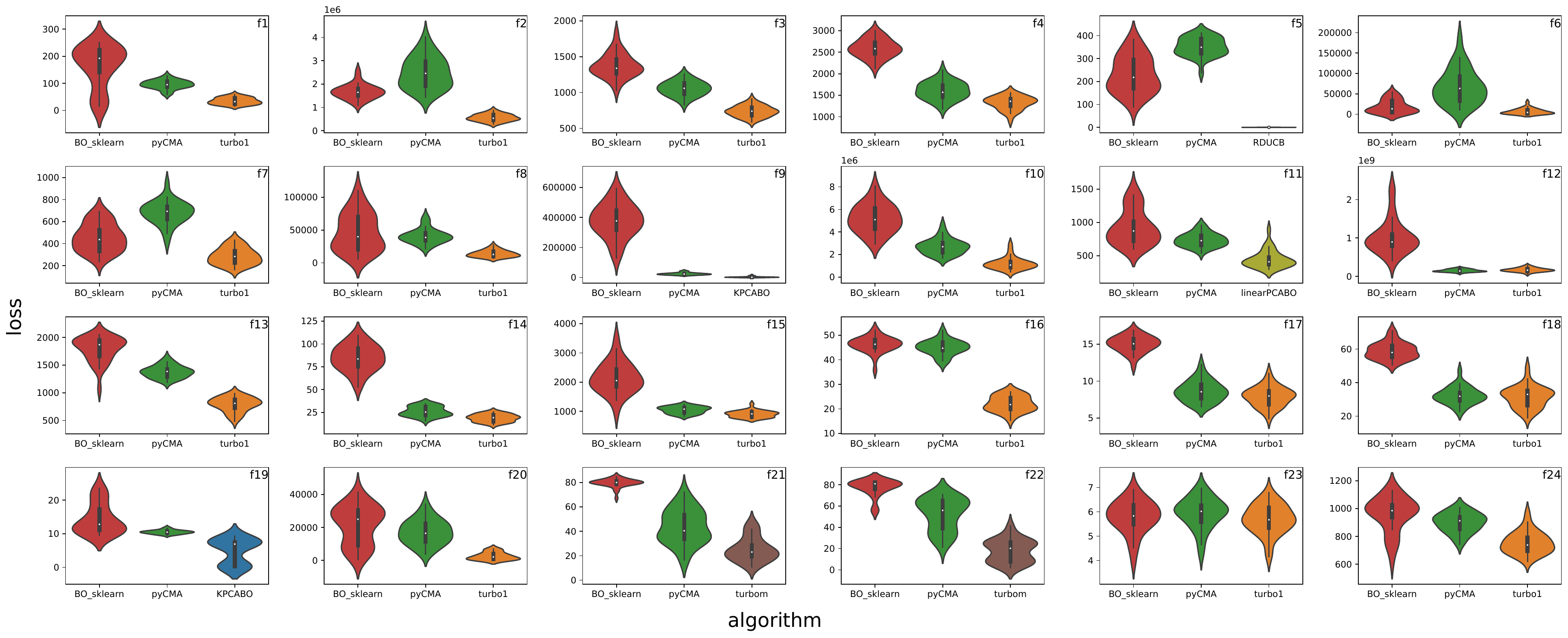}
      \caption{The violin plots show, for dimension 60 and budget 650 (final budget), the distribution of the best-so-far function values for all functions, obtained by vanilla BO, CMA-ES, and the best among the HDBO algorithms. The plots include a marker for the median of the data and a box indicating the interquartile range.}
    \label{boxplotD60}
\end{figure}

Figures~\ref{boxplotD10},~\ref{boxplotD20}, and~\ref{boxplotD60} provide a comparison through violin plots for the performance of standard BO, CMA-ES, and the best among the HDBO algorithms for each setting (combination of function and dimension) at the end of the budget. Here we show results for dimensions 10, 20, and 60, while dimension 40 was already discussed in Section~\ref{sec:furtherdiscussion}.
\label{sec:moreviolin}

\section{ECDF Analysis}
\label{sec:ecdf}
For a more comprehensive analysis of the results, we also present cumulative distribution function (ECDF) plots. These plots were generated in the `Fixed-Target Results: Cumulative Distribution' section of IOHanalyzer~\cite{IOHprofiler}. They illustrate the aggregated ECDF curves for selected target values across all 24 BBOB functions. The term `aggregated' refers to a comprehensive overview resulting from the collective performance across the entire range of functions. For each function, we select 10 target values logarithmically distributed within the range defined by the minimum and maximum values attained by all the algorithms. For a given number of random targets and each method, our plots show the percentage (shown on the y-axis) of runs that achieved that target within a given budget (shown on the x-axis), with this percentage then averaged across all the different fixed targets.
\subsection{Dimension D = 10}
\begin{figure}[ht]
\center
\includegraphics[width=0.8\columnwidth]{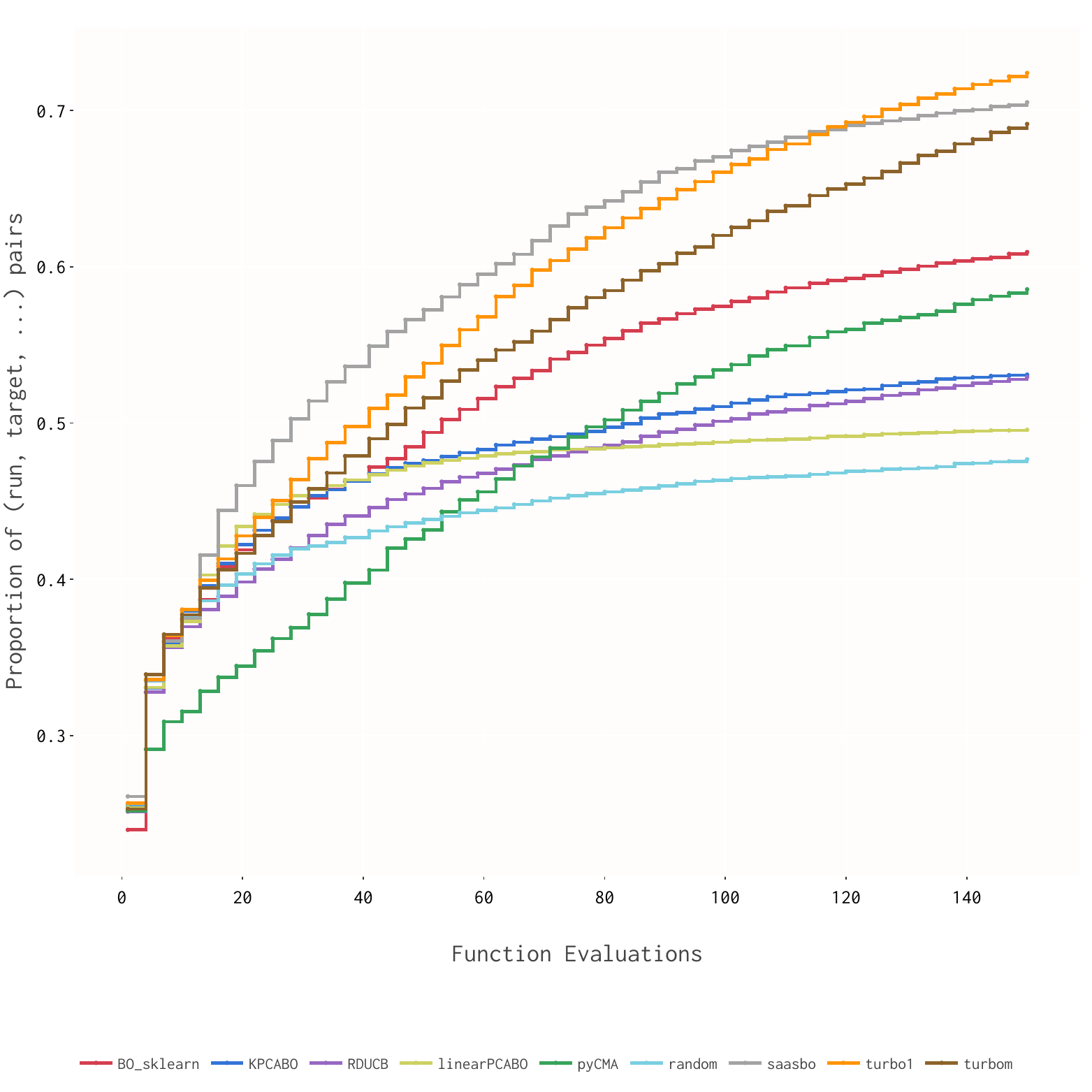}
      \caption{`Fixed-Target Results: Cumulative Distribution': Aggregated ECDF curves for selected target values for dimension 10.}
    \label{ECDF_10D}
\end{figure}
Figure~\ref{ECDF_10D} confirms the assertions made in Section~\ref{sec:10D}. Random search proves to be the least effective method. Initially, CMA-ES exhibits poor performance, but it shows a slight improvement over time. 
Vanilla BO demonstrates strong performance and greater robustness across evaluations when compared to other algorithms like PCA-BO, KPCA-BO, RDUCB, and CMA-ES. SAASBO achieves the highest percentage of target values nearly throughout the entire budget range. Towards the end of the run, TuRBO1 continues to enhance the percentage without stagnation, overcoming SAASBO at approximately 110 evaluations.
\subsection{Dimension D = 20}
\begin{figure}[ht]
\center
\includegraphics[width=0.8\columnwidth]{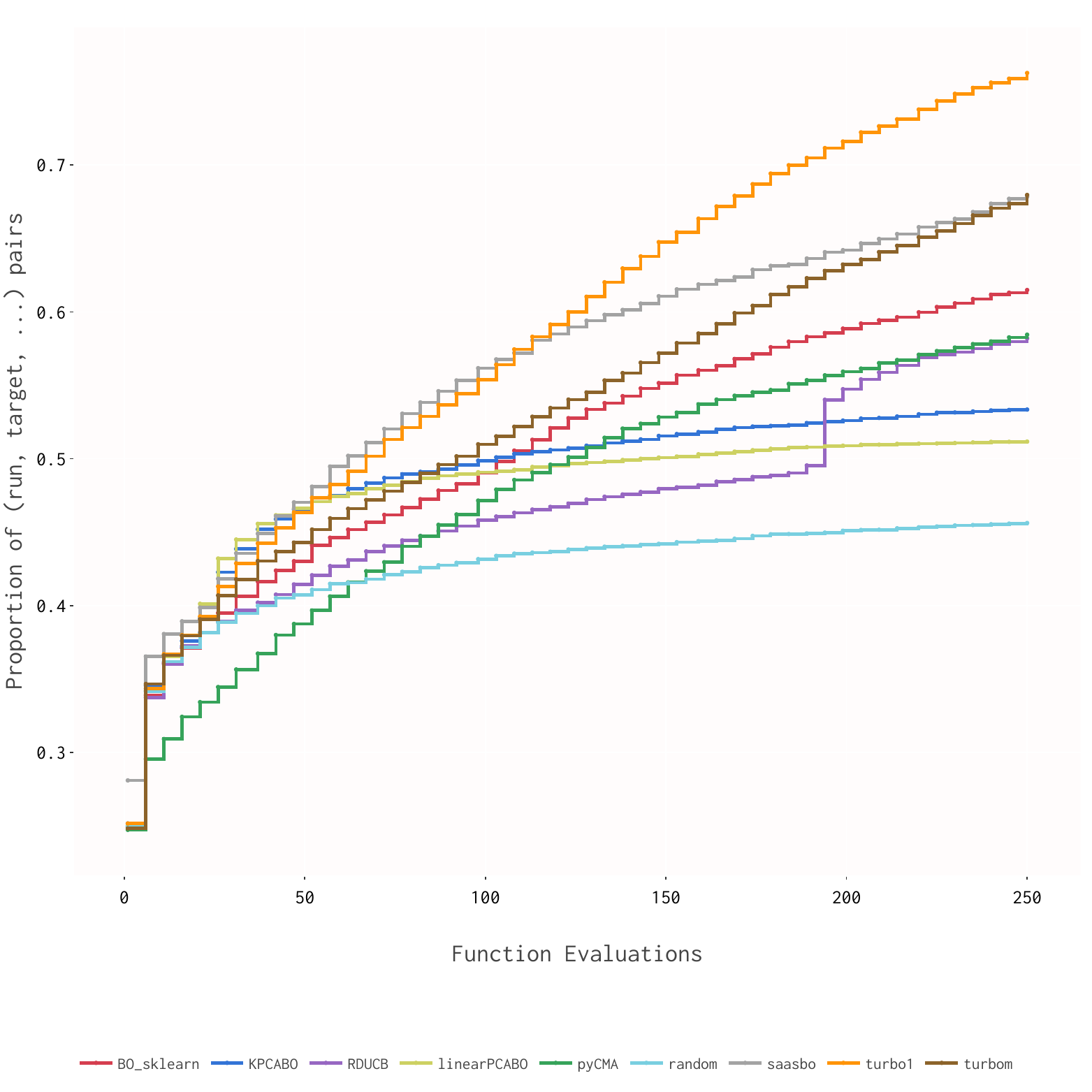}
      \caption{`Fixed-Target Results: Cumulative Distribution': Aggregated ECDF curves for selected target values for dimension 20.}
    \label{ECDF_20D}
\end{figure}
Figure~\ref{ECDF_20D} validates the claims stated in Section~\ref{sec:20D} and distinctly illustrates that all the examined algorithms outperform random search. Even in dimension 20, vanilla BO clearly outperforms CMA-ES.
SAASBO and TuRBO1 exhibit comparable performance within a small evaluation budget.
As the number of evaluations increases, TuRBO1 demonstrates more pronounced distinctions compared to other algorithms than in dimension 10, surpassing SAASBO. It is important to note that, for this dimension, we have complete data only for f15-f24. Hence, aggregation is based only on these functions. The sudden performance boost of RDUCB explained in Section~\ref{sec:20D}, is also evident: around budget 200, all runs show an improvement, increasing the probability of reaching the target value.
\subsection{Dimension D = 40}
\begin{figure}[ht]
\center
\includegraphics[width=0.8\columnwidth]{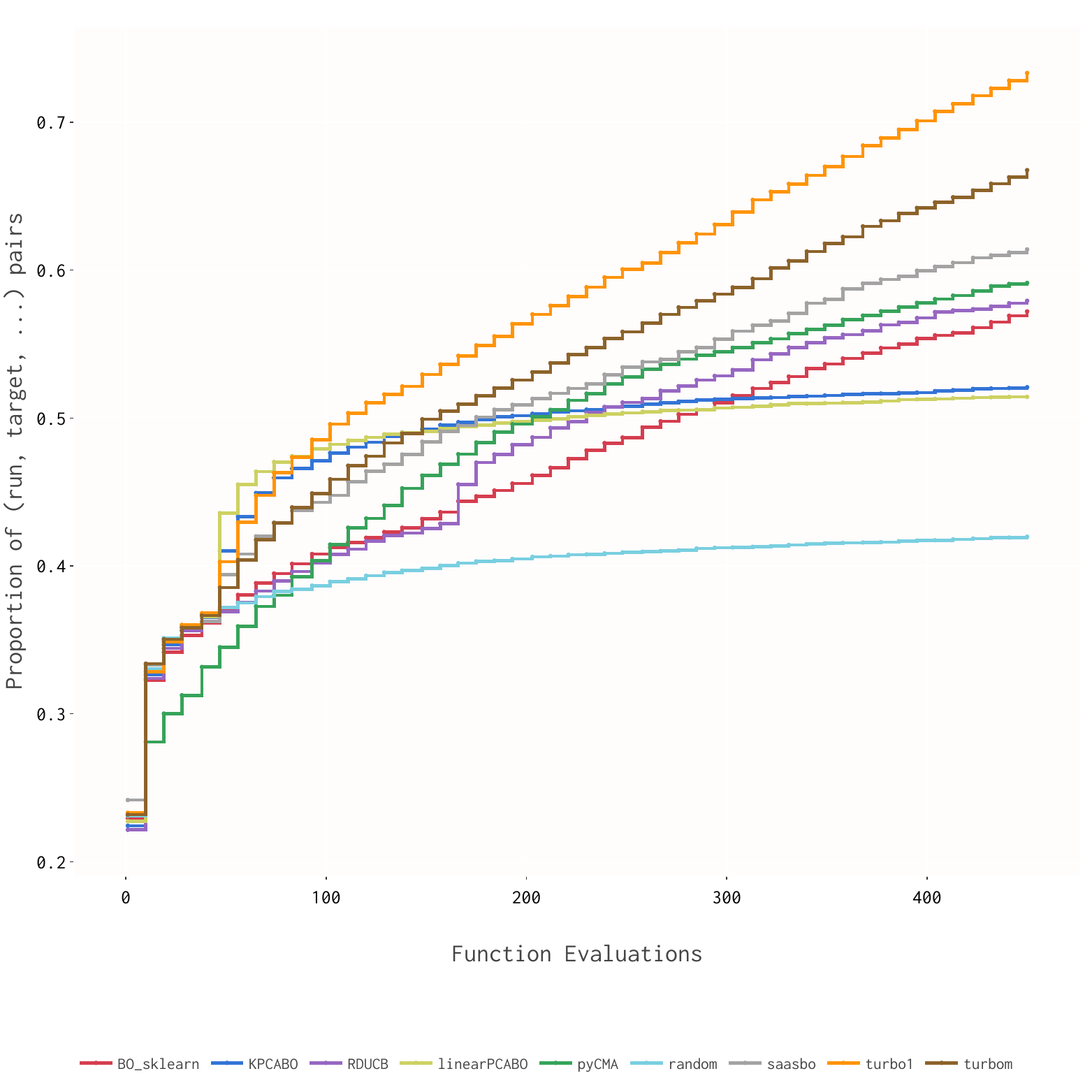}
      \caption{`Fixed-Target Results: Cumulative Distribution': Aggregated ECDF curves for selected target values for dimension 40.}
    \label{ECDF_40D}
\end{figure}
Figure~\ref{ECDF_40D} is in line with the analysis presented in Section~\ref{sec:40D}.
It is worth noting that while vanilla BO performs strongly up to dimension 20, here it stands out as one of the less effective strategies. It is also noteworthy that PCA-BO shows a slightly higher probability of reaching the target values than other algorithms for small budgets.
As the budget increases, the superiority of TuRBO1 over all other algorithms is evident. SAASBO is no longer comparable with TuRBO1, even if the aggregation is based only on the functions f15-f24, as in dimension 20.
\subsection{Dimension D = 60}
\begin{figure}[ht]
\center
\includegraphics[width=0.8\columnwidth]{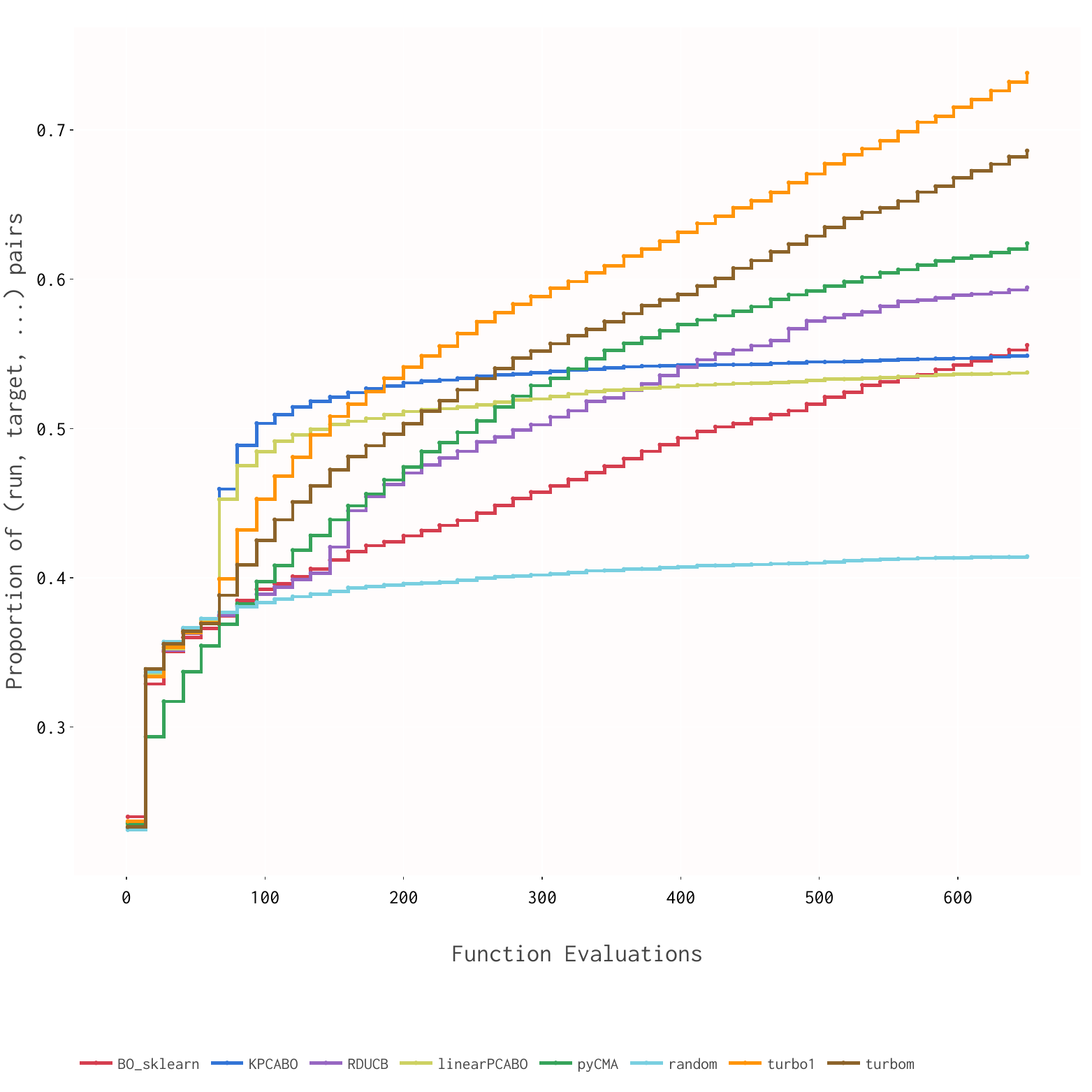}
      \caption{`Fixed-Target Results: Cumulative Distribution': Aggregated ECDF curves for selected target values for dimension 60.}
    \label{ECDF_60D}
\end{figure}
The ECDF curves in Figure~\ref{ECDF_60D} validate the observations presented in Section~\ref{sec:60D}. The ECDF curve for SAASBO is unavailable for dimension 60, as experiments for this algorithm were not completed due to the prohibitive computational costs of the runs. BO clearly suffers from the high dimension. The potential of PCA-BO and KPCA-BO for small evaluation budgets becomes even more evident than in dimension 40. They stand out as the most likely to reach the target values, which highlights the effectiveness of KPCA-BO and PCA-BO in the early stages of optimization. However, they tend to plateau around evaluation 200. On the other hand, TuRBO demonstrates an enhancement in the percentage with an increasing budget, positioning it as one of the most effective methods. Considering the full evaluation budget, it is followed by CMA-ES and RDUCB.
\end{document}